\documentclass{article}

\usepackage{arxiv}
\usepackage[title]{appendix}

\usepackage[utf8]{inputenc} % allow utf-8 input
\usepackage[T1]{fontenc}    % use 8-bit T1 fonts
\usepackage{hyperref}       % hyperlinks
\usepackage{url}            % simple URL typesetting
\usepackage{booktabs}       % professional-quality tables
\usepackage{amsfonts}       % blackboard math symbols
\usepackage{nicefrac}       % compact symbols for 1/2, etc.
\usepackage{microtype}      % microtypography
\usepackage{graphicx}
\usepackage{amsmath,amsthm,amsfonts,amscd} 
\numberwithin{equation}{section}
\usepackage{float}
\usepackage{graphicx}
\graphicspath{{figures/}}
\usepackage{graphics}
\usepackage{eucal}
\usepackage{verbatim}
\usepackage{subcaption}
\usepackage{makeidx}
\PassOptionsToPackage{hyphens}{url}\usepackage{hyperref}
\usepackage{draftcopy}

\usepackage{rotating}
\usepackage{afterpage}
\usepackage[ruled,vlined,resetcount,algosection]{algorithm2e}
\usepackage{multirow}
\usepackage{epsfig}
\DeclareMathOperator*{\argmin}{arg\,min}

\title{Latent Space Arc Therapy Optimization}

\author{
 Noah Bice$^1$, Mohamad Fakhreddine$^1$, Ruiqi Li$^1$, Dan Nguyen$^2$, Christopher Kabat$^1$, \\
 \textbf{Pamela Myers$^1$, Niko Papanikolaou$^1$, and Neil Kirby$^1$} \\
  $^1$Department of Radiation Oncology, UT Health San Antonio, San Antonio, TX \\
  $^2$Medical Artificial Intelligence and Automation (MAIA) Laboratory, Department of \\
  Radiation Oncology, UT Southwestern Medical Center, Dallas, TX\\
  \texttt{noah.bice@outlook.com} \\
}

\begin{document}
\maketitle
\begin{abstract}
Volumetric modulated arc therapy planning is a challenging problem in high-dimensional, non-convex optimization. Traditionally, heuristics such as fluence-map-optimization-informed segment initialization use locally optimal solutions to begin the search of the full arc therapy plan space from a reasonable starting point. These routines facilitate arc therapy optimization such that clinically satisfactory radiation treatment plans can be created in about 10 minutes. However, current optimization algorithms favor solutions near their initialization point and are slower than necessary due to plan overparameterization. In this work, arc therapy overparameterization is addressed by reducing the effective dimension of treatment plans with unsupervised deep learning. An optimization engine is then built based on low-dimensional arc representations which facilitates faster planning times.
\end{abstract}
% \keywords{Arc Therapy \and deep Learning \and optimization}

%%%%%%%%%%%%%%%%%%%%%%%%%%%%%%%%%%%%%%%%%%%%%%%%%%%%%%%%%%%%%%%%%%%%%%%%%%
\section{Introduction}\label{intro}

External beam radiotherapy (EBRT) is a very common modality in cancer treatment. In EBRT, patients rest on a treatment couch and are treated with a photon beam generated by a linear accelerator (linac). External beam therapy is ideal for both individual patients and entire hospital systems, since it is non-invasive, effective, and accommodates a relatively high patient throughput. One major consideration for EBRT treatment planning is photon attenuation in healthy tissues. The usual physical model for narrow beam photon attenuation is to suggest that the rate of change of photon fluence $\phi$ with respect to depth $x$ is directly proportional to the fluence at that depth, i.e.
\begin{equation*}
\begin{gathered}
    \frac{d\phi}{dx} = - \mu \phi(x) \ , \\ 
    \implies \phi(x) = \phi_0 e^{-\mu x} \ .
\end{gathered}
\end{equation*}
This model must be modified to describe the dose deposition of clinical radiotherapy beams, but the overall rapid attenuation with depth still applies \cite{khan2014khan}. To treat a deep target with external beam photon therapy, some radiation dose will be deposited in surrounding tissues. The physical limitations of radiation dose deposition motivate the practice of radiation treatment planning, during which customized radiation beams are manufactured to treat the target and minimize the risk of toxicities.

Radiation treatment planning has developed substantially over the course of a few decades. 3D conformal radiotherapy, a practice in which the target is simply cut out with collimators in the beam's eye view and treated, was considered a state of the art technique only three decades ago. In the 1990's, intensity modulated radiotherapy (IMRT) was designed under Brahme, Webb, Bortfeld, and Boyer, amongst others \cite{webb2015intensity}. The development of IMRT marked a leap in radiotherapy technology. In IMRT treatment planning, 3D CRT beam shapes are decomposed into smaller beamlets (or bixels), and the intensity of those beamlets are mathematically optimized to suit the patients' dose distribution needs. Optimized fluence profiles are recreated with multileaf collimators (MLC) in the head of the linac, which use many narrow tungsten blocks (``leaves'') to create smooth edges and match structure boundaries in the beam's eye view.

\begin{figure}[!hbt]
\centering
\includegraphics[width=0.6\linewidth]{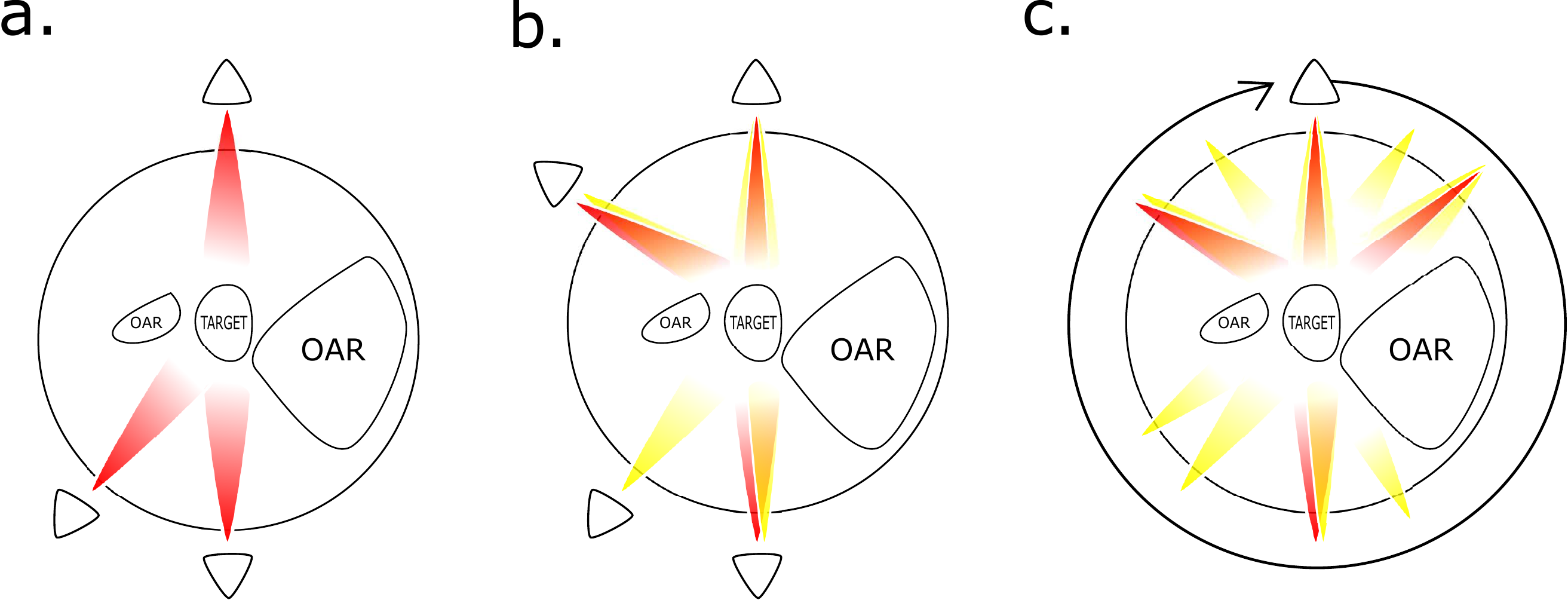}
\caption[Arc Therapy Description]{\textbf{a)} 3D Conformal therapy uses uniform fluences at discrete gantry angles to treat targets at depth. \textbf{b)} In IMRT, modulated beam profiles are delivered with a series of apertures at each angle. \textbf{c)} VMAT delivers a series of co-planar beams in a single sweep of the linac's gantry.}
\label{vmatabc}
\end{figure}

Volumetric modulated arc therapy (VMAT) is an extension of IMRT. In VMAT, the treatment planner asserts that delivery should be performed continuously over an arc. Rather than creating a dose distribution based on a few discrete gantry angles, up to 180 co-planar beams are used to describe the radiation delivery of a sweeping arc (Figure \ref{vmatabc}c).

The requirement that the beams for a single arc be co-planar and deliverable with one sweep of the gantry\footnote{In practice, two or three arcs with different collimator angles is common.} encourages rapid treatments, and the use of so many beams promotes target coverage and normal tissue sparing \cite{otto2008volumetric, teoh2011volumetric}. The practice of arc therapy is supported by many clinical trials and retrospective analyses \cite{wolff2009volumetric, davidson2011assessing, tsai2011treatment}. Despite the many clinical benefits of VMAT over conventional IMRT, the treatment planning software must manage significantly more computational complexity.

\subsection{Optimization Formalism}\label{convoptrout}

VMAT plans are specified by beam weights and leaf positions for about 160 leaves for around 180 control points throughout a delivery, depending on the linear accelerator and treatment planning preferences. The task of VMAT plan creation is then an optimization problem with order-10,000 variables whose objective is defined by complicated radiation transport. A formal statement of the VMAT optimization problem was laid out in detail by Unkelbach \textit{et al.} in 2015 \cite{unkelbach}. In this formulation, VMAT plans are parameterized as a collection of discrete apertures distributed over the arc. We use the array $x^\phi_{ij}$ to represent the transmission of beamlets $ij$ at gantry angle $\phi$, where each beamlet represents a small rectangular area in the beam's eye view. For a specified machine and patient geometry, we can numerically evaluate the dose at voxel $k$ due to one monitor unit (MU) through beamlet $ij$ and store the results in a dose influence matrix $D^\phi_{ijk}$. During the optimization procedure, the dose calculation for voxel $k$ is reduced to a simple summation, 
\begin{equation}\label{dosesum}
d_k  = \sum_\phi \sum_{ij} D^\phi_{ijk} x^\phi_{ij} \ .
\end{equation}
In the context of direct aperture optimization (DAO), apertures are specified by the position of MLC leaves on the left and right banks of the collimator \cite{shepard2002direct, earl2003inverse}. If we take the second index of $x$ to be the leaf indexing direction, we can write 
\begin{equation}\label{apdef}
x^\phi_{ij} = y^\phi [\Theta(i - l^\phi_j) - \Theta(i - r^\phi_j)]] \ ,
\end{equation}
where $\Theta$ is the Heaviside step function, $y^\phi$ is the beam weighting (MUs delivered), and $r^\phi_j$ and $l^\phi_j$ are the right and left leaf end index positions for leaf $j$ at control point $\phi$. This is to say that beamlet $ij$ is set to $y^\phi$ if it falls between $l^\phi_j$ and $r^\phi_j$ and zero otherwise.

In this case, we can write out the full optimization problem as such: for dosimetric objective function $f(d)$, return
\begin{equation}
 y^*, l^*, r^* =  \argmin_{y, l, r}  f(d; y, l, r)
\end{equation}
subject to 
\begin{equation}
d_k  = \sum_\phi \sum_{ij} D^\phi_{ijk} x^\phi_{ij}
\end{equation}
\begin{equation}
x^\phi_{ij} = y^\phi [H(i - l^\phi_j) - H(i - r^\phi_j)]
\end{equation}
\begin{equation}
y^\phi \geq 0
\end{equation}
\begin{equation}\label{nonabutting}
l^\phi_j \leq r^\phi_j
\end{equation}
The last two constraints require that aperture intensities are all non-negative and MLC leaves are non-abutting. Notice also that the resolution of the beamlets in the leaf-indexing direction is limited by the leaf width.

\subsubsection{Traditional Approaches to Arc Therapy Optimization}\label{tradsection}

There are two popular approaches to VMAT optimization which have been implemented clinically. These algorithms solve VMAT with satisfactory accuracy but are often slower than desired. An outline of these algorithms is provided below.

\textit{Global DAO with geometry-based segment initialization:} This algorithm was pioneered by Earl \textit{et al.} in 2003 and largely remains unchanged in modern clinical implementations \cite{earl2003inverse}. MLC leaf positions are initialized based on the shape of targets and OARs in the beam's eye view. Following initialization, some stochastic search algorithm such as simulated annealing (SA) or a genetic algorithm is used to search the full VMAT plan space \cite{cotrutz2003segment}. Stochastic algorithms such as these are constructed to allow navigation of local optima, so a poor initialization does not significantly harm the accuracy of the solution. The solution state is iteratively adjusted until the dose constraints are reasonably satisfied. This optimization structure is implemented in Varian's RapidArc \cite{vanetti2011role}.

\begin{figure}[!hbt]
\centering
\includegraphics[width=0.5\linewidth]{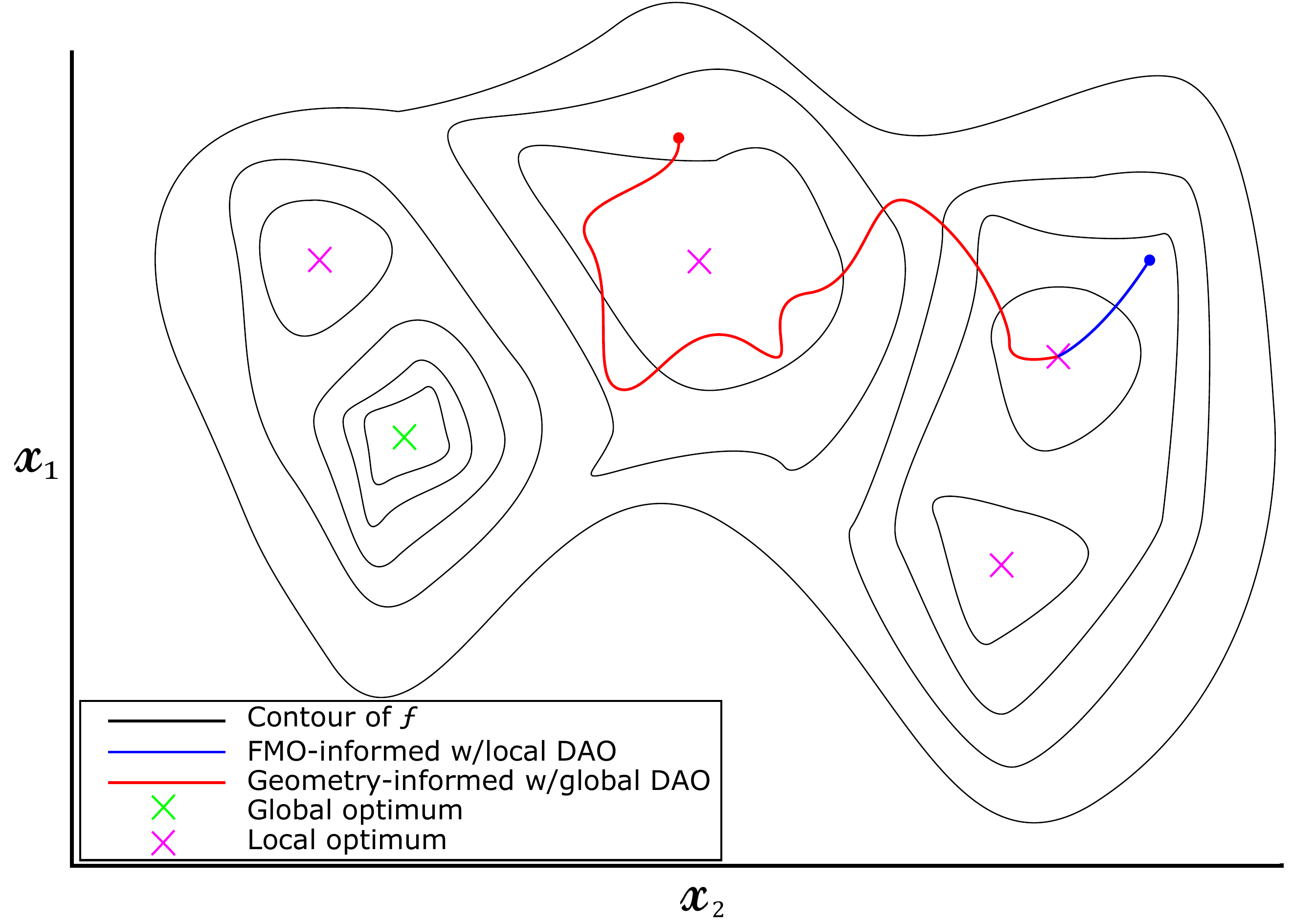}
\caption[VMAT Optimization with Traditional Algorithms]{Global DAO with geometry-informed segment initialization uses a simple initialization which is refined with stochastic DAO. Local DAO with FMO-informed segment initialization uses a decent initialization with optimized fluences at individual control points, then refines the solution with a gradient-based search. Notice that neither of these algorithms are expected to achieve the true global optimum.}
\label{traditionalalgs}
\end{figure}

\textit{Local DAO with FMO-informed segment initialization:} Bzdusek \textit{et al.} developed this approach to introduce gradient information in the VMAT search process \cite{bzdusek2009development}. The initialization step in this algorithm is inspired by conventional IMRT optimization. Apertures at each control point are individually optimized with convex, gradient-based fluence map optimization (FMO). Following the individual optimizations, some arc sequencing procedure is used to represent the optimized fluences with physically deliverable MLC positions and beam weights. The overall solution state at this point is expected to be at a decent accuracy, from which gradient-based DAO can take over. Variations of this algorithm have been implemented in treatment planning systems such as SmartArc (Philips), Oncentra VMAT (Nucletron), RayArc (RaySearch Laboratories), and Monaco (Elekta) \cite{unkelbach, feygelman2010initial}. These two algorithms are represented visually in Figure \ref{traditionalalgs}.

Despite their desirable tumor coverage and normal tissue sparing, VMAT plans are problematically slow to create. Performing a single iteration of the VMAT planning process with modern computing equipment can take up to 10 minutes, and creating an acceptable plan often requires multiple iterations of dosimetrists adjusting clinical objectives \cite{tian2015multi}. Because of this time constraint, VMAT planning is a major clinical bottleneck. Additionally, creating multiple treatment plans per patient, as is required in adaptive radiotherapy, is not logistically feasible (at least on a large scale). The average radiotherapy patient must have their VMAT plan created in advance, and therefore incurs the risk of overdosing healthy tissue while underdosing the target. We suspect that VMAT creation is slow for the same reason it is so successful; the variety of possible treatment plans is so vast that searching for an appropriate plan is almost prohibitive. 

\subsection{Unsupervised Learning and VAEs}

The purpose of unsupervised learning algorithms is to identify and exploit structure within data. The textbook example of unsupervised machine learning is principal component analysis (PCA). In PCA, the directions of maximum variation within the training data are identified with singular value decomposition \cite{wold1987principal}. In 1991, Turk and Pentland famously developed a primitive facial recognition software based on PCA \cite{turk1991eigenfaces}. In their work, they identify the first $N$ principal components of a facial image dataset and use those images to define basis vectors in a low-dimensional ``face space.'' Rather than characterizing new facial images with the regular pixel representation, images are projected into the principal subspace for comparison. With this technique, low-level information such as image noise texture is generally ignored, and distinct images of the same person's face will have similar face space coordinates following their projection.  Notice that this method, as opposed to supervised techniques, does not require the person which is being identified by the software to exist within the training dataset. PCA learns the ``interesting'' directions in the high-dimensional image space which can be applied to new images at any time.

\begin{figure}[!hbt]
\centering
\begin{subfigure}{.49\textwidth}
  \centering
  \includegraphics[width=\linewidth]{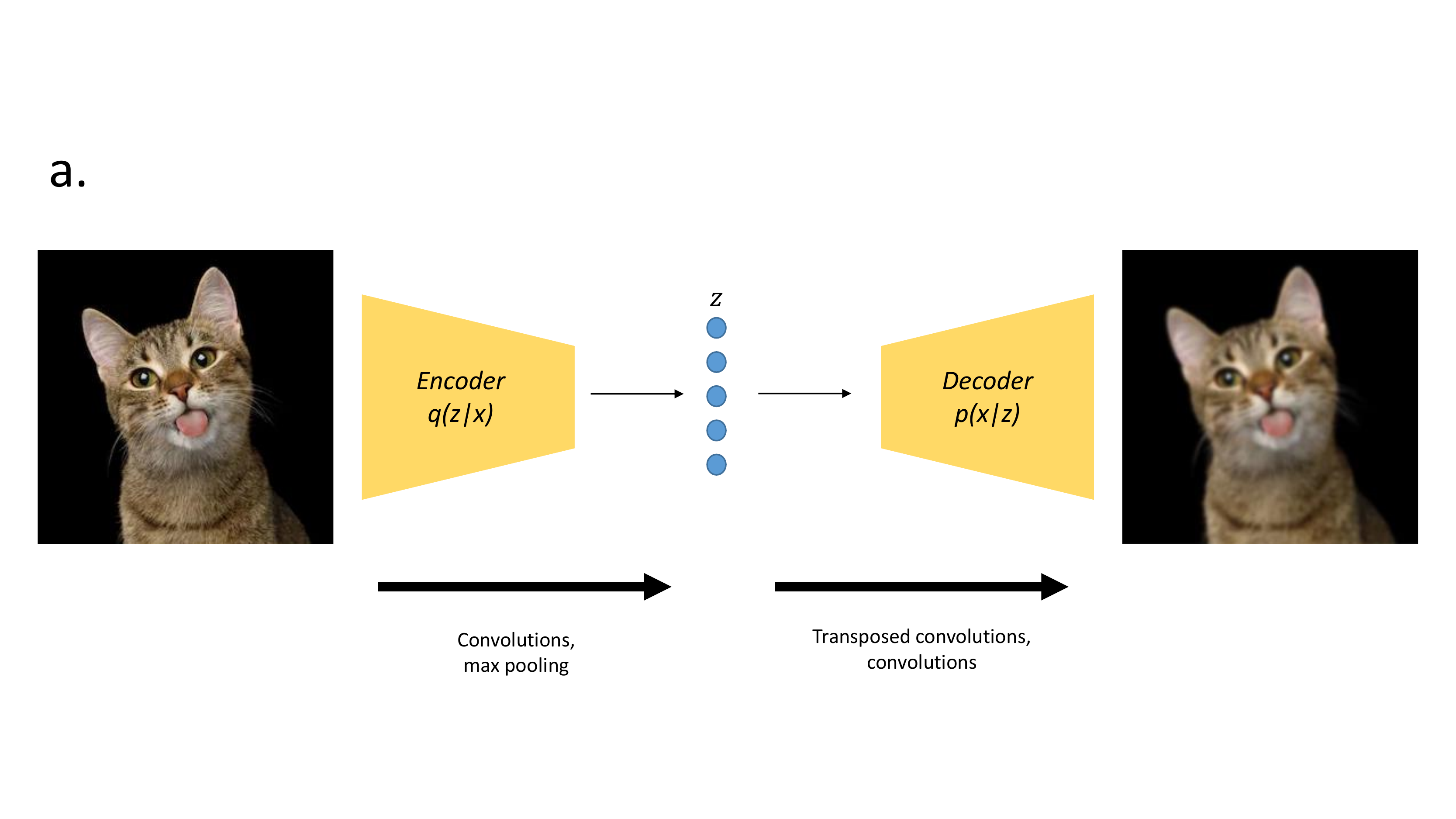}
\end{subfigure}
\begin{subfigure}{.49\textwidth}
  \centering
  \includegraphics[width=\linewidth]{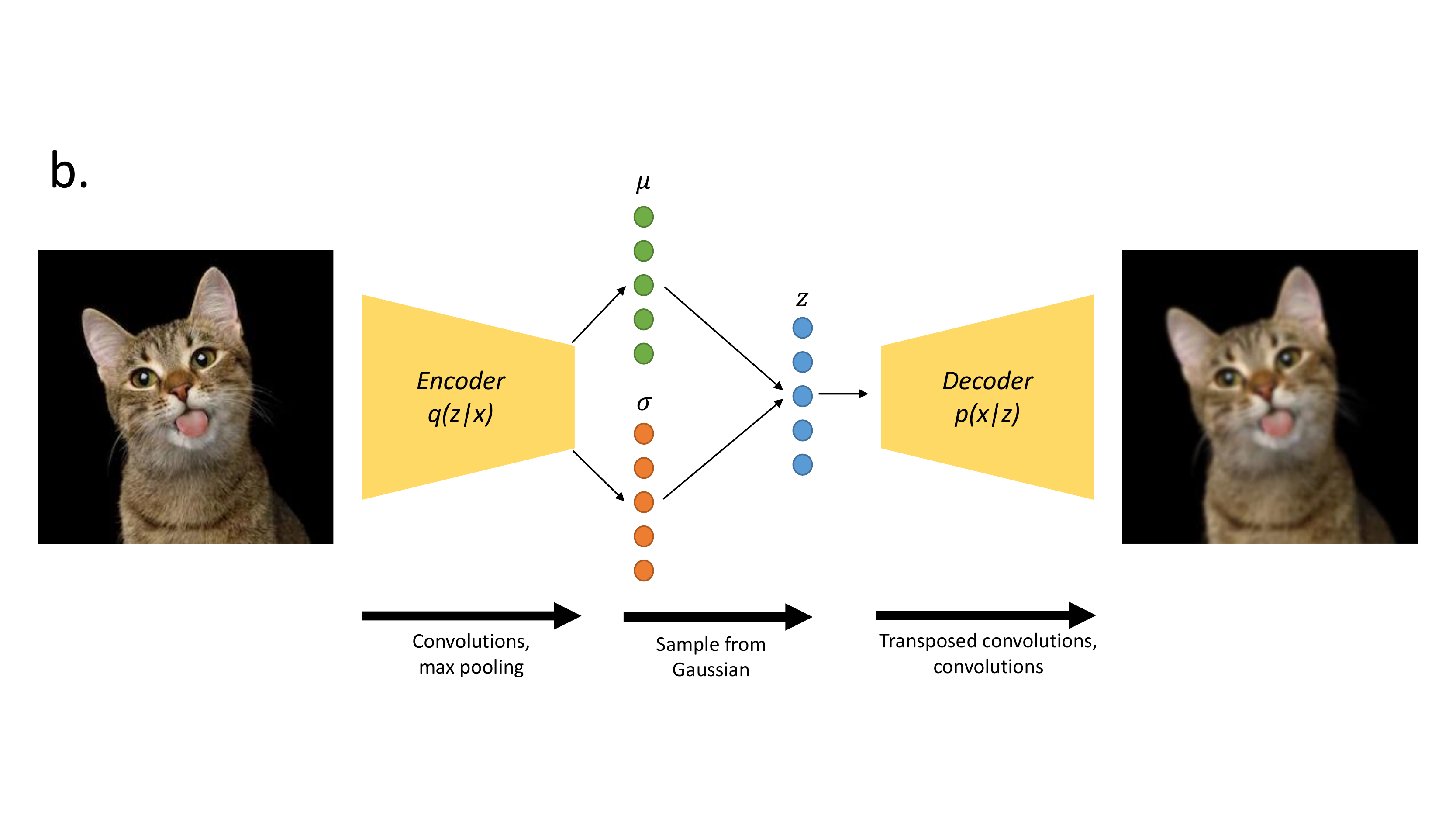}
\end{subfigure}
\caption[Variational Autoencoders]{Autoencoders are a standard deep learning model for dimensionality reduction. \textbf{a)} An encoder and decoder are used to compress the image and reconstruct it. \textbf{b)} Variational autoencoders map images to a ``cloud'' in the latent space, specified by coordinates and spreads. VAEs generally have better sampling properties for image generation than traditional AEs.}
\label{catsvae}
\end{figure}

The canonical example of unsupervised learning with neural networks is deep autoencoders (AE) \cite{goodfellow2016deep}. Similar to PCA, the objective of training an autoencoder is to learn an efficient method of representing high-dimensional data with low-dimensional coordinates. The task required of AEs is to compress the training data into a specified number of dimensions and reconstruct them accurately (Figure \ref{catsvae}a). Formally, we require that model weights minimize reconstruction loss which is often implemented as the mean squared error between training examples $x^{(j)}$ and their reconstructions $\hat{x}^{(j)}(w) = \mbox{AE}(x^{(j)}; w)$,
\begin{equation}
    L(w) = \frac{1}{2N} \sum_{j=1}^N \sum_{i} (x_i^{(j)} - \hat{x}_i^{(j)}(w))^2 \ .
\end{equation}
Intuitively, AEs are required to learn the most efficient way to encode the data they are shown. Autoencoders' mechanisms for compression are usually fit to suit the training data; for example, image AEs might use many convolutional layers, while natural language AEs might use recurrent units.

One utility of unsupervised learning is for generative modeling. Once an autoencoder has learned the directions of significant variation in the data space, latent variables can be randomly generated and transformed into convincing examples in the input space. While conventional autoencoders are practically easy to construct and train, they are often inadequate for generating new high-dimensional data. For these AEs,  we simply expect the models to learn generalization implicitly by showing them a high volume of training examples; there is no requirement for the models to organize the latent space efficiently. Consequentially, two input data points which we might consider semantically similar, such as two images of the same person or cat, might be mapped to very different positions in the latent space with no effect on the overall loss function. Certain properties which are good for image synthesis, such as the ability to interpolate between points in the low dimensional space, are not ensured in the conventional autoencoder framework. These issues are however addressed by variational autoencoders (VAE).

VAEs are structurally similar to conventional AEs but introduce an extra step following compression. Rather than mapping each input to a specific point, images are mapped to entire regions of the low-dimensional space. For an $n$-dimensional latent space, each compressed data point is specified by an $n$-dimensional Gaussian having $n$ means and standard deviations. When training the model, the Gaussians are sampled from to find some low-dimensional point which is used to create a reconstructed image. This process ensures that the space between the mapped means is semantically meaningful and might be used to generate new, realistic images \cite{kingma2013auto}. 

Additionally, during VAE training we require that the compressed data are organized nicely; they are not crowded in one region of the latent space with the rest of the space being semantically meaningless. This is encouraged by using a Kullbach-Leibler (KL) divergence term in the loss function \cite{kullback1951information}. The KL divergence is used to measure the dissimilarity between a distribution of compressed inputs and an $n$-dimensional Gaussian centered at zero. For compressed data $\{Q^{(j)}\}$, the VAE loss can be written
\begin{equation}\label{vaelossfn}
    L(w) = \bigg[\frac{1}{2N} \sum_{j=1}^N \sum_{i} (x_i^{(j)} - \hat{x}_i^{(j)}(w))^2\bigg] + \alpha \bigg[\frac{1}{2} \sum_{k=1}^{dim(Q)} \log \sigma_{Q_k}^2 - \sigma_{Q_k}^2 - \mu_{Q_k}^2 + 1\bigg] \ ,
\end{equation}
where $\alpha$ is a hyperparameter determining the relative contribution of the KL term.

Besides AEs and VAEs there are many other popular deep unsupervised learning methods which can be used for dimensionality reduction. One very common method is with generative adversarial networks (GAN) \cite{goodfellow2014generative}. GANs, however, are famously unstable during training, so for this work we focus on autoencoders exclusively \cite{kodali2017convergence}. Beyond GANs, there are several other mechanisms for dimensionality reduction including flows and hybrid approaches such as adversarial autoencoders \cite{kingma2018glow, makhzani2015adversarial}. For future implementations of this work, we expect that these models might better describe the practical VMAT subspace and are worth exploring. 

\subsection{Arc Therapy and the Curse of Dimensionality}

In previous implementations of arc therapy optimization the problem is addressed in the fully-parameterized arc planning space. This is problematic because the size of the search space for any optimization problem grows with the power of the number of optimization parameters. For VMAT optimization, a typical optimizer might be required to select 80 leaf positions on 2 MLC banks over the duration of 80 control points distributed throughout the arc. This yields $80*2*80=12,800$ optimization parameters. If $N$ discrete positions are allowed for every leaf, the optimizer must navigate a space of $N^{12,800}$ unique VMAT plans. Of course, modern VMAT algorithms are more sophisticated than a brute force search of $100^{12,800} = 10^{25,600}$ values, or whatever $N$ is allowed. Clever initializations and gradient-based searches such as those in the FMO-informed algorithm provide solutions in relatively satisfactory accuracy and time. Yet the problem of effectively navigating the solution space persists; gradient-based methods are confined to the region in the neighborhood of their initialization point while global stochastic methods are inefficient in searching such a large space.

Rather than suggesting a better initialization or search routine in the fully-parameterized VMAT space, we would like to simplify the problem altogether with pattern-based dimensionality reduction. In the proposed framework, we reduce the effective dimension of arc therapy data prior to plan optimization to facilitate a more computationally efficient search. In doing so, the optimizer works in a space with inherently fewer local optima and is less susceptible to trapping. Additionally, the volume of the learned space is significantly smaller, only permitting $N^d$ possible solutions for latent space dimension $d$. In this setting, the optimizer is limited to only consider regions of the full space where VMAT plans have existed historically.

\begin{figure}[!hbt]
\centering
\includegraphics[width=0.35\linewidth]{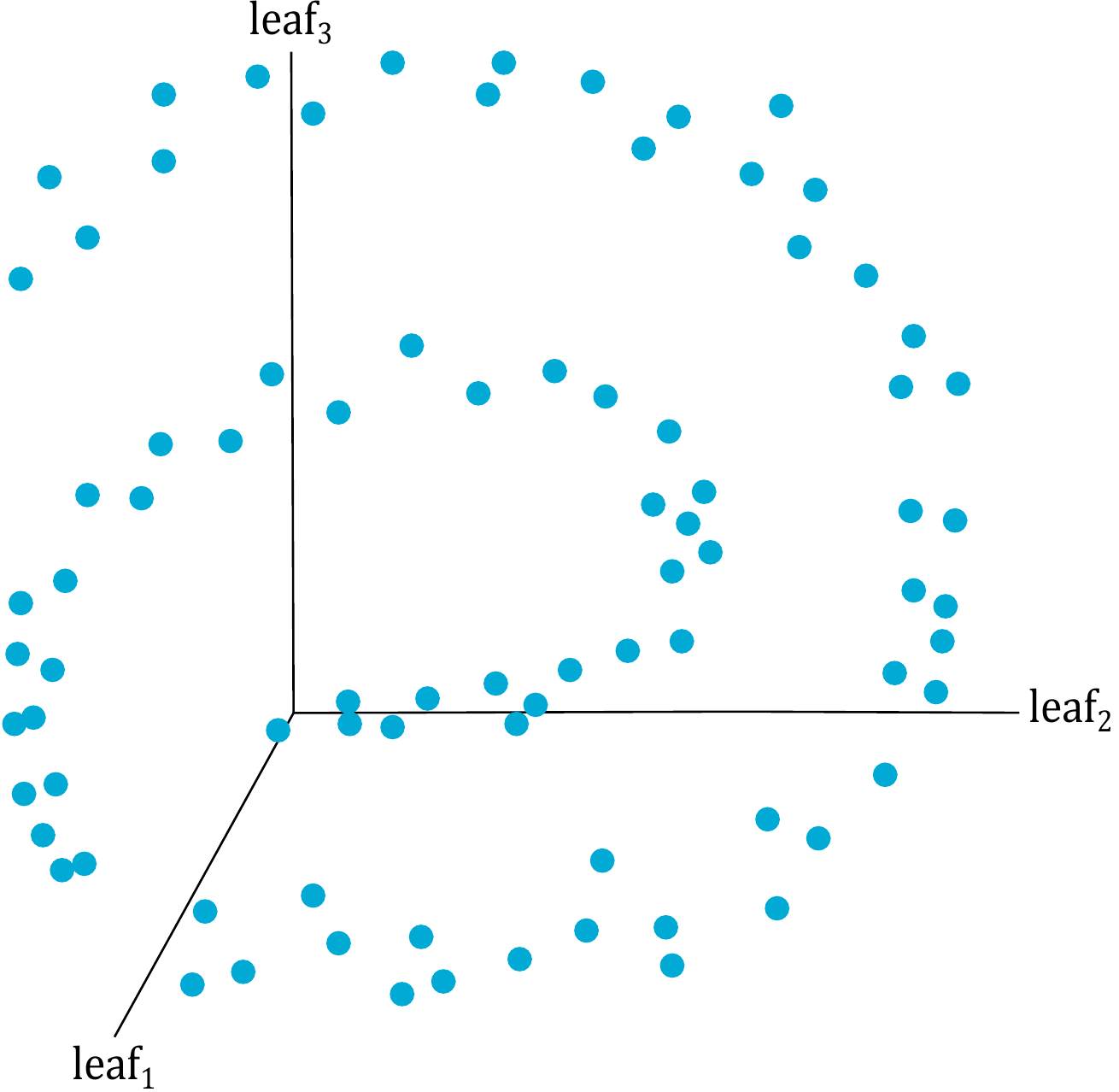}
\caption[Arc Manifold]{Here, the raw VMAT data is symbolized with points described by three coordinates. While these values may be necessary for the physical description of machine parameters, all of the interesting variation happens along one direction specified by a spiral. We conjecture that historical arc therapy data exists on a connected surface like this, whose coordinates we might learn with an autoencoder algorithm.}
\label{manifold}
\end{figure}

When talking about AEs, and particularly VAEs, it is common to describe the high-dimensional data generating distribution as a manifold\footnote{Although the term ``manifold'' has a precise definition in mathematics, it is used rather loosely in the context of machine learning. However, the central concept of the space being \textit{connected} is shared between the two communities.}. While it is useful to suggest that VMAT data are simply overparameterized, it carries much more weight to postulate that the data populate a small, connected region of the VMAT plan space which is characteristic of a manifold (Figure \ref{manifold}).

This is regularly an implicit hypothesis when using any sort of machine learning dimensionality reduction. The proposition that arc therapy solutions inhabit a low-dimensional manifold embedded within the space that we use describe them appears to be a novelty of this work. We cite two sources of evidence for this conjecture.

First, arc data is densely distributed in the full planning space. Sampling a random point in 12,800-dimensional space and finding a clinically reasonable arc is no more likely than finding an image of a cat\footnote{We have no extra information about the relative size of the cat and arc therapy subspaces, but we imagine they are comparable.}. The region inhabited by clinically viable VMAT arcs represents only a small portion of the total solution space. This is shown in Figure \ref{datasetfig}.

Second, we expect that the space between example plans is occupied by other reasonable plans. Additionally, we anticipate that VMAT solutions for alike disease sites and anatomies will be similarly connected. For example, the bladder-empty and bladder-full regions of planning space should be connected with a range of reasonable bladder-half-full plans. This particular property is exploited by VAEs, where the boundaries of compressed training examples are blurred by the stochastic encoding procedure.

\subsection{Latent Space Optimization in Literature}

\begin{figure}[!hbt]
\centering
\includegraphics[width=0.6\linewidth]{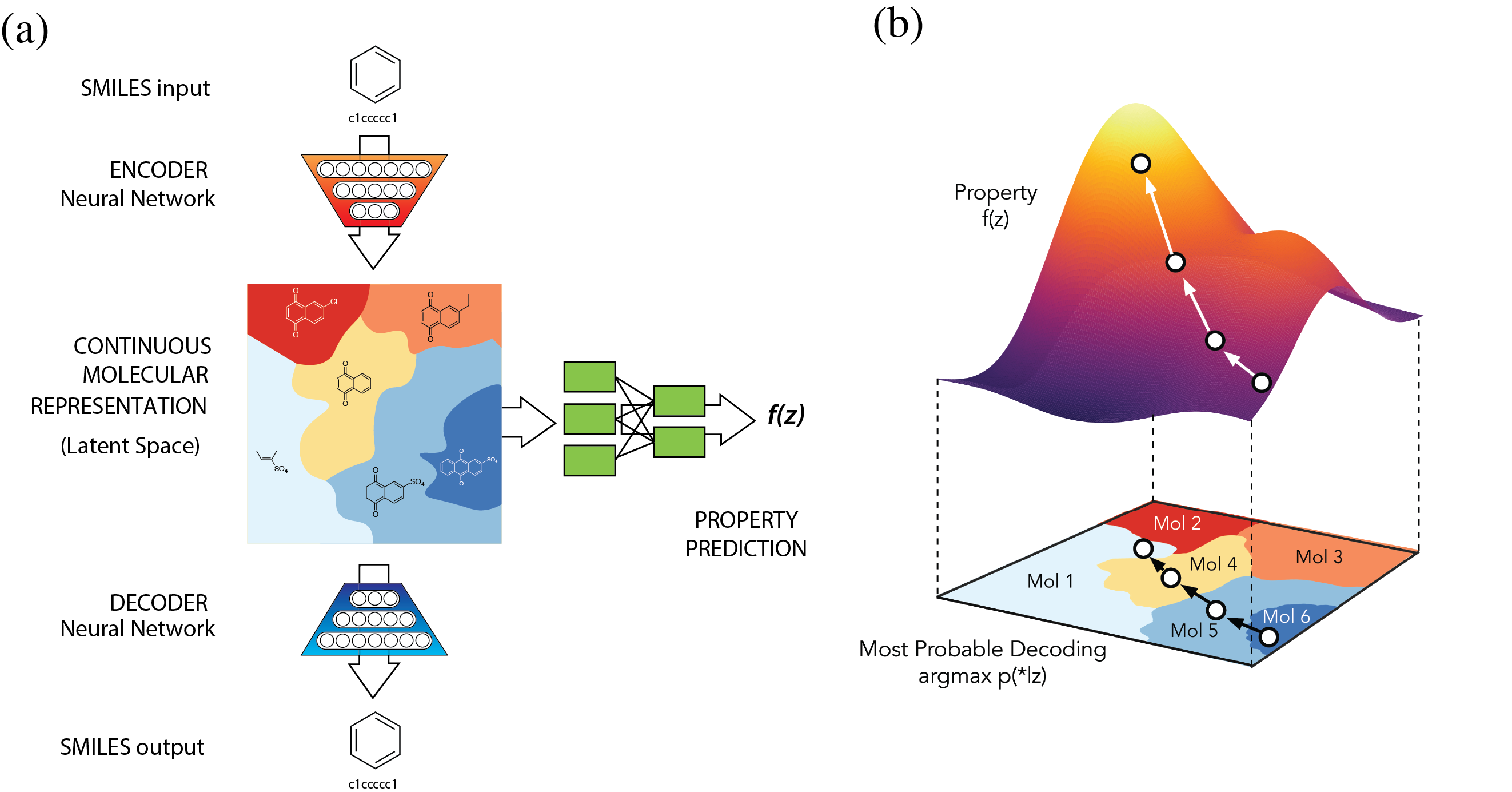}
\caption[Drug Discovery with AEs]{Gomez-Bombarelli \textit{et al.} suggest the drug discovery framework pictured here. \textbf{a)} A deep autoencoder is used to establish a low-dimensional continuous representation of molecules. \textbf{b)} The latent space can be navigated with an optimization algorithm to satisfy some objective $f$. Figure reproduced with permission from \cite{gomez2018automatic}.}
\label{drugdiscovery}
\end{figure}

The use of compressed representations of data often reduces computational and memory requirements. One relevant use of VAEs by Gomez-Bombarelli \textit{et al.} was to facilitate drug discovery \cite{gomez2018automatic}. In their framework, discrete representations of molecules are compressed using a deep AE, and the learned continuous representations are used to craft molecules with desirable properties (Figure \ref{drugdiscovery}). This technique has also been used to facilitate synthetic gene design, which faces the same problems of high-dimensionality and non-convexity as drug synthesis and VMAT optimization \cite{gonzalez2015bayesian}.

Latent space optimization has also been studied in more formal settings such as in the works of Lu \textit{et al.}, Moriconi \textit{et al.} and Raponi \textit{et al.} \cite{lu2018structured, moriconi2019high, raponi2020high}. It has been shown that optimization within low-dimensional learned spaces can outperform traditional techniques, especially for very high-dimensional problems.

\section{Arc Therapy Dataset}\label{datasetsection}

To reduce the effective dimension of VMAT plans we describe them based on patterns in historical clinical deliveries. Patterns emerge in arc therapy MLC positions primarily due to three influences:

\begin{enumerate}
    \item Adjacent MLC leaves are often at similar positions. Apertures often conform to target and organ at risk (OAR) boundaries in the beam's eye view, with neighboring leaves only slightly displaced.
    \item Arc sequencing requires that MLC leaf positions do not vary drastically from one control point to the next. The same MLC leaves are required to be at similar positions for small time steps throughout the delivery. 
    \item Treatments of the same disease site are expected to have similar deliveries. Similarities in anatomy between patients will lead to similar collimation strategies for different patients.
\end{enumerate}

A dataset of 1,874 unique arc therapy deliveries was recovered from Elekta MLC log files. All treatments were performed on Elekta Versa HDs at UT Health’s Mays Cancer Center between 2014 and 2020. These data contained MLC leaf positions (mm) and instantaneous dose rates (MU/s) recorded at 25 Hz during treatment. No personal patient data are recorded in these files. Each arc was resampled to represent 80 evenly spaced control points over the course of the arc, and instantaneous dose rates were numerically integrated to give the cumulative monitor units (MU) delivered for each control point. Since the Versa HD has two MLC leaf banks with 80 leaves each, every plan was described with 2 * 80 * 80 = 12,800 parameters. For consistency within the dataset, only arcs with total arc length $>$ 300 degrees were retained.

\begin{figure}[!hbt]
  \centering
  \includegraphics[width=0.85\linewidth]{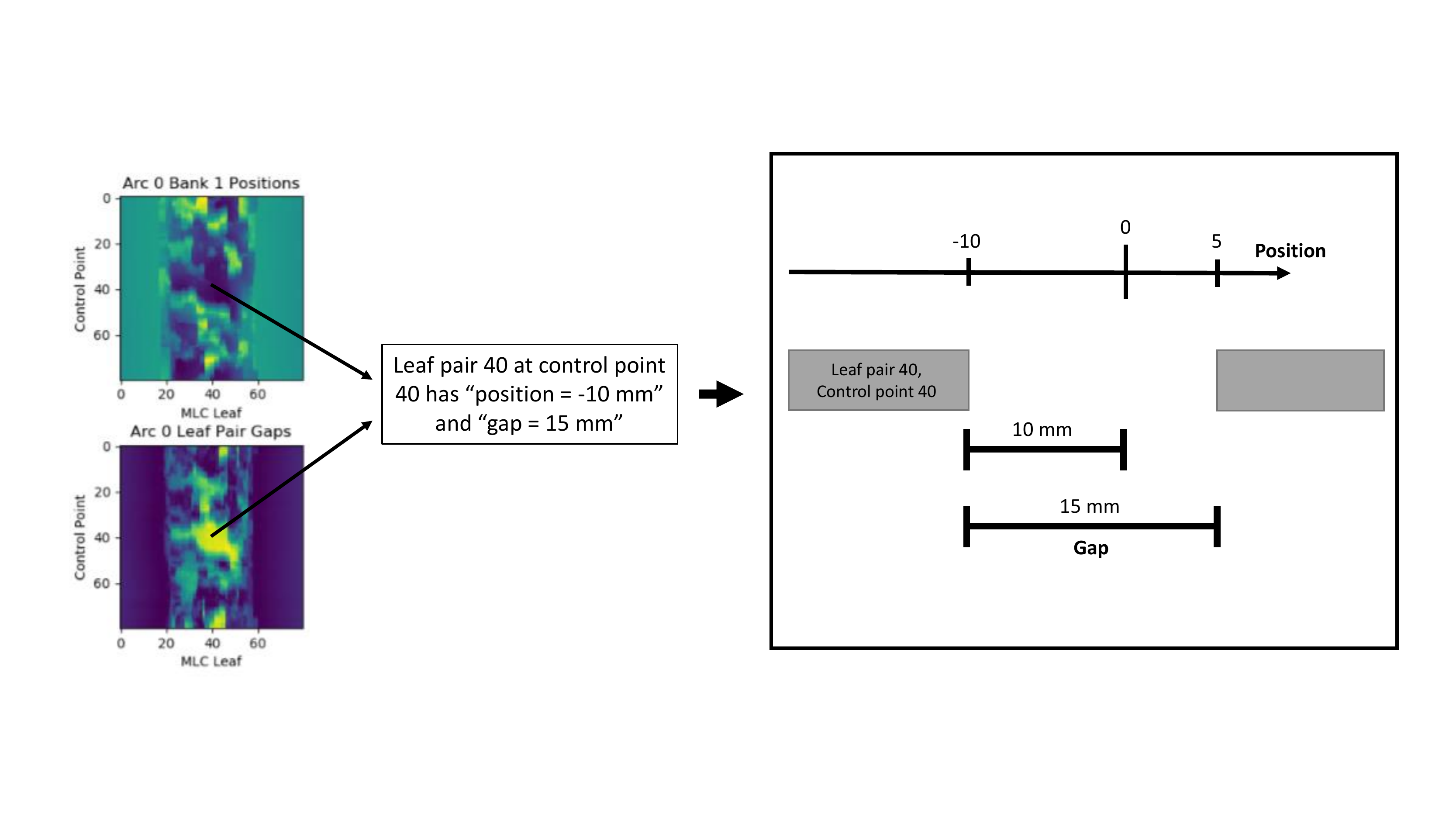}
\caption[Image Coordinates]{The leaf position content of full VMAT plans is entirely described in a two-channel image. The first channel is used to describe the leaf-end position of one bank while the other is parameterized by the distance between leaf ends for the two banks.}
\label{imagecoords}
\end{figure}

To ensure that the deep learning frameworks only make physically realizable predictions, rather than describing MLC leaf pairs with two positions, leaf pairs were described with one leaf end position and with the gap between leaf ends for the leaf pair. With this system, we can satisfy the requirement that leaf ends are non-abutting (Equation \ref{nonabutting}) by requiring that predicted gap values always be non-negative (Figure \ref{imagecoords}). This can easily be implemented with deep learning models using rectified linear unit (ReLU) activations in the output layer. 

\begin{figure}[!hbt]
  \centering
  \includegraphics[width=0.65\linewidth]{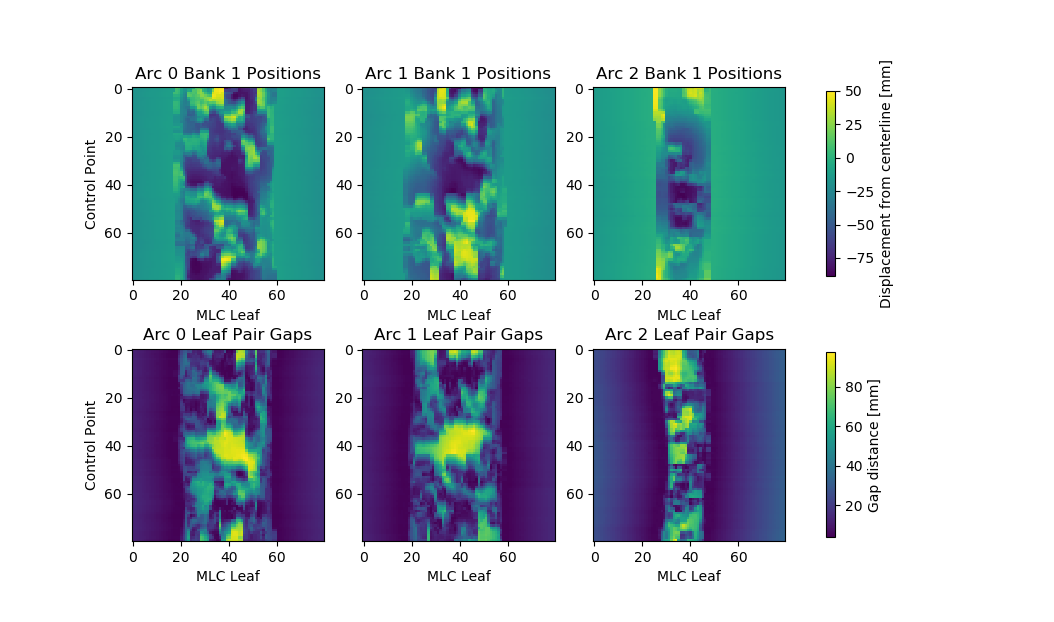}
\caption[Arc Therapy Dataset]{Elekta log file MLC leaf positions with each arc resampled to 80 control points. Leaf positions and gap distances are shown for 5 arcs from the training dataset. In the top row, positions of leaves on the first bank are parameterized in distance from the leaf end to the centerline. In the bottom row, the opposing leaves are parametrized in terms of their leaf end distance between leaves. Notice that the minimum value in the bottom row of images is zero. }
\label{datasetfig}
\end{figure}

\begin{figure}[!hbt]
\centering
\includegraphics[width=0.6\linewidth]{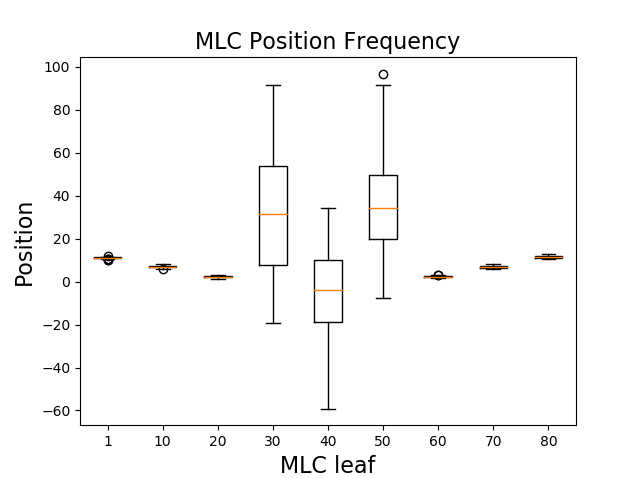}
\caption[MLC Leaf Position Frequencies During VMAT]{The positional frequency for some MLC leaves during one VMAT plan (Figure \ref{datasetfig} Arc 2) are visualized. Notice that some leaves vary over the duration of treatment more significantly than others.}
\label{mlcfrequency}
\end{figure}

At this point, the VMAT plans can be visualized as two-channel 80 x 80 images (Figure \ref{datasetfig}). Each channel describes the leaf end positions of one bank of leaves over the duration of the treatment. Some patterns in the data are visible in Figure \ref{datasetfig}. Notice that yellow and blue pixels show up in patches; this is a direct consequence of 1) and 2) above. Additionally, notice that in most cases not all the leaves move drastically over the course of treatment. In arc 2, for example, only leaves between about 30 and 50 appear to move at all during treatment (Figure \ref{mlcfrequency}). This is simple but resonating motivation to consider some machine learning dimensionality reduction.

\subsection{Preprocessing}

\begin{figure}[!hbt]
\centering
\includegraphics[width=0.8\linewidth]{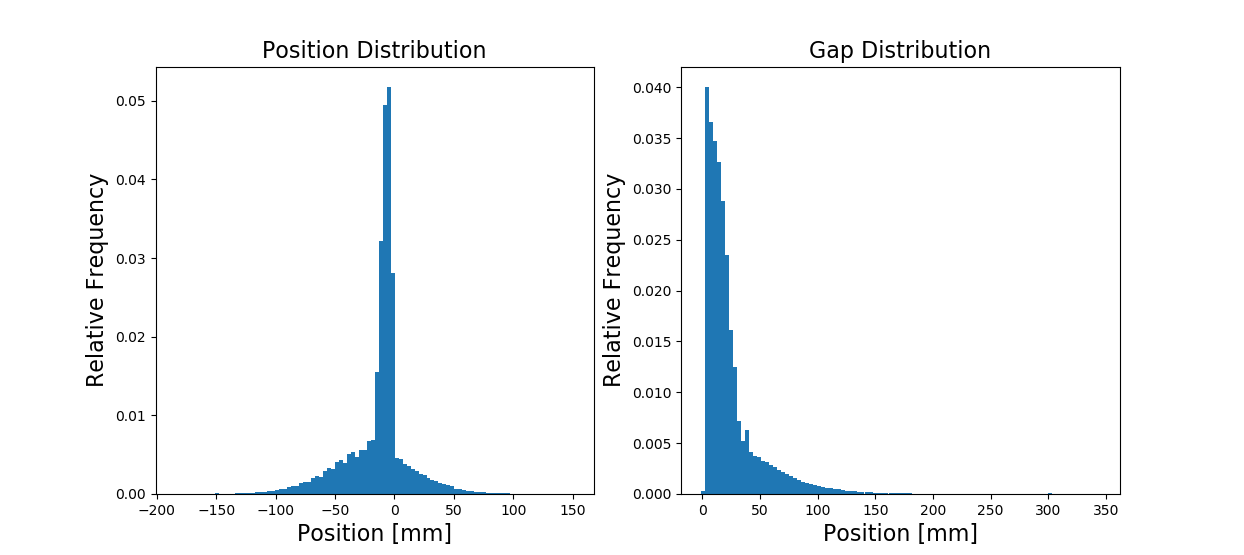}
\caption[Distribution of Arc Data by Channel]{The distribution of arc data in the position-gap parameterization. The median absolution position is 10.5 mm and the median gap distance is 16.8 mm. These values are used to inform the normalization procedure.}
\label{channeldist}
\end{figure}

Because standard values for positions and gaps are significantly different, we normalize the data differently for individual channels. The distributions of leaf positions and gaps are shown in Figure \ref{channeldist}. The median value of entries in the gap channel is significantly higher than the median for the position channel. However, much of the interesting variation in the data happens for small gap values. Manually determined normalization values of 8 times the channel median were used to scale the data, with 84.0 mm chosen for leaf positions and 132.8 mm for leaf pair gaps. To prepare the data for input into neural networks they were normalized and randomly split into a training set of 1,686 examples (90\%) and test set of 188 examples (10\%).

%%%%%%%%%%%%%%%%%%%%%%%%%%%%%%%%%%%%%%%%%%%%%%%%%%%%%%%%%%%%%
\section{Learning the VMAT Subspace}\label{ch3}

In this section, we aim to identify a machine learning model which best compresses VMAT plans. Using the collected arc therapy dataset, several lossy compression algorithms are trained to reduce the effective dimension of the data. We consider standard PCA alongside conventional AEs and VAEs. The deep learning models use successive convolutional layers and max pooling for feature extraction in the arc dataset. Validation arcs are compressed and reconstruction accuracy is reported. We visualize the reconstructed arcs and compare model performances with the median absolute reconstruction error measured in mm.

\subsection{Principal Component Analysis}

\begin{figure}[!hbt]
\begin{subfigure}{0.49\textwidth}
  \centering
  \includegraphics[width=\linewidth]{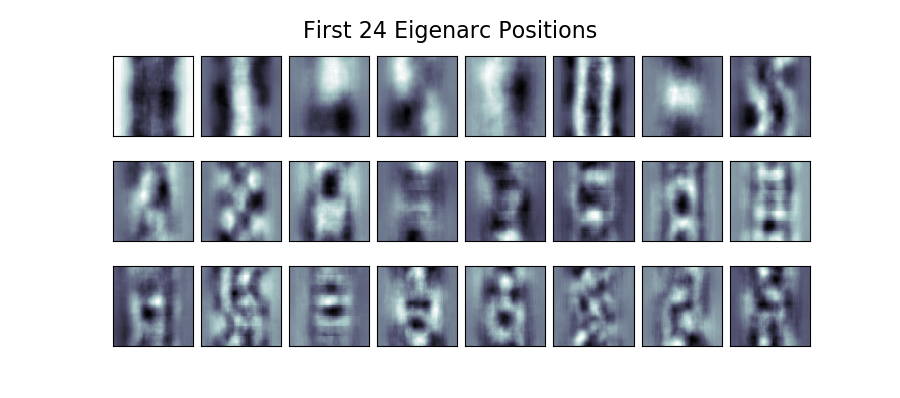}
\end{subfigure}
\begin{subfigure}{0.49\textwidth}
  \centering
  \includegraphics[width=\linewidth]{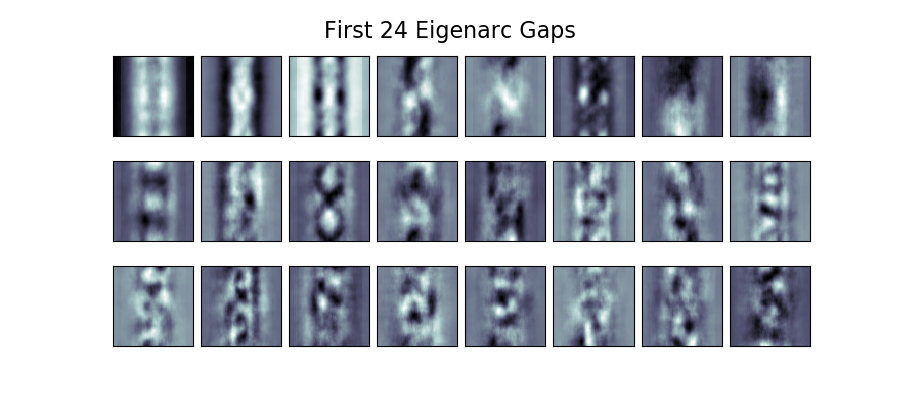}
\end{subfigure}
\caption[First 24 Eigenarcs]{First 24 principal components of arc dataset (sorted from left to right and top to bottom). These images show the directions of maximum variation in the arc therapy planning space.}
\label{pcafig}
\end{figure}

For comparison of the deep learning models with a conventional unsupervised learning technique, dimensionality reduction with PCA was used to compress the arc data. This method is identical to the one used by Turk and Pentland for their facial recognition work \cite{turk1991eigenfaces}. Similar to the the principal components of the facial image dataset which they coined ``eigenfaces,'' we can visualize the first 24 principal components of the VMAT dataset (Figure \ref{pcafig}). These ``eigenarcs'' are the regular modes of arc therapy delivery and define an orthogonal basis for the subspace learned with PCA. Reconstructions are produced as a linear combination of the learned basis vectors visualized in Figure \ref{pcafig}.

\begin{figure}[!hbt]
\centering
\begin{subfigure}{0.49\textwidth}
  \centering
  \includegraphics[width=\linewidth]{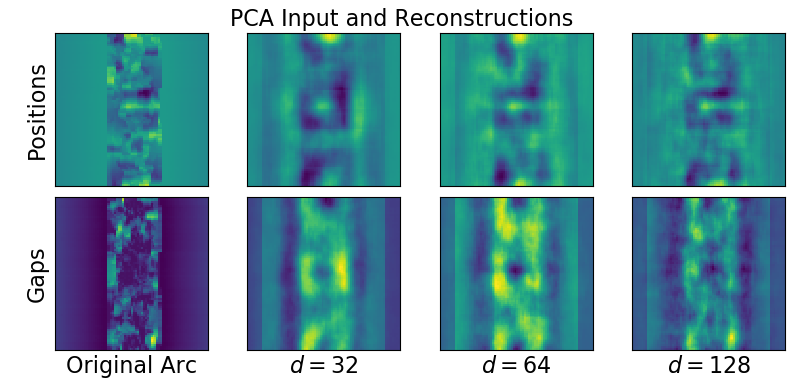}
\end{subfigure}
\begin{subfigure}{0.49\textwidth}
  \centering
  \includegraphics[width=\linewidth]{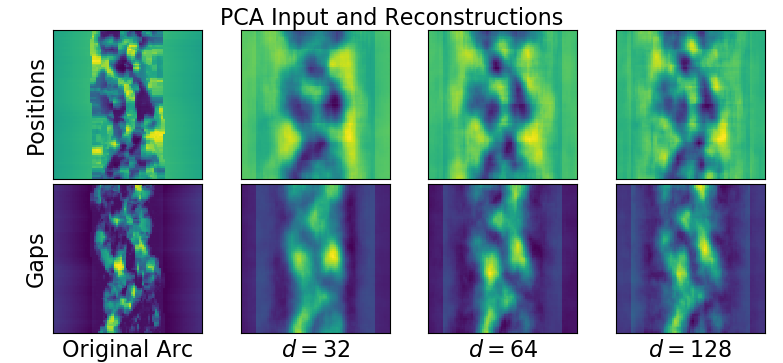}
\end{subfigure}
\caption[PCA Reconstructions]{Validation arc reconstructions with PCA. These reconstructions based on a linear machine learning model are analogous to the VAE reconstructions shown in Figure \ref{vaereconstructions}.}
\label{pcareconstructions}
\end{figure}

\begin{figure}[!hbt]\centering
   \begin{minipage}{0.48\textwidth}
     \centering
     \includegraphics[width=\linewidth]{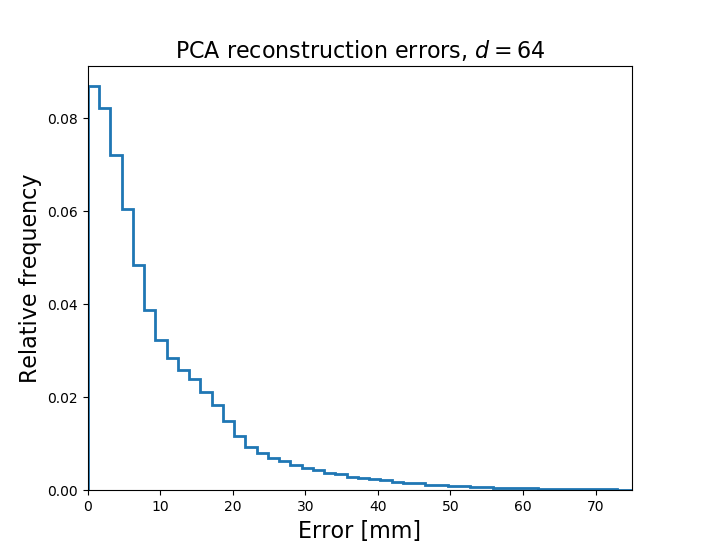}
     \caption[PCA Reconstruction Error Distribution]{An analogue of Figure \ref{reconerrors} is shown for the $d=64$ PCA compressions. For this specific model, we observe a median absolute reconstruction error of 6.82 mm across the validation dataset.}
     \label{pcaerrordist}
   \end{minipage}
    \hfill
   \begin {minipage}{0.48\textwidth}
     \centering
     \includegraphics[width=\linewidth]{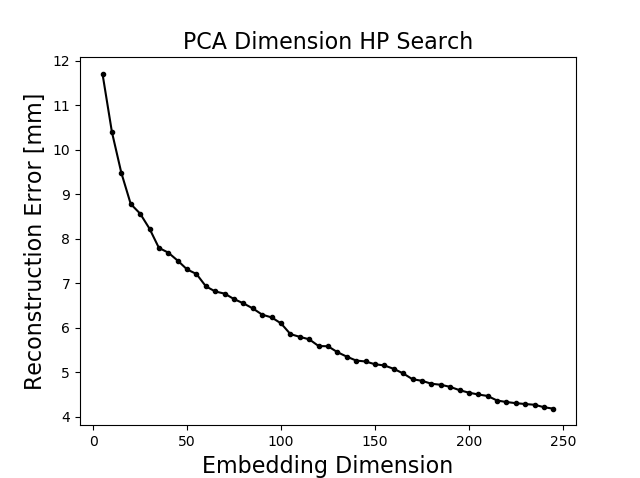}
     \caption[PCA Dimension HP Search]{The median absolute reconstruction error across all validation arcs is shown for embedding dimensions from 5 to 245 using PCA. We notice an ``elbow'' somewhere between 32 and 128 dimensions.}
    \label{pcadimensionfig}
   \end{minipage}
\end{figure}

Absolute leaf reconstruction errors $|\hat{x} - x|$ for PCA can be evaluated and visualized as in Figure \ref{pcaerrordist}. The distribution is heavily skewed towards higher values and suggests that metrics such as the root mean squared error (RMSE) might give a biased representation of typical error for a given leaf. With a 64-dimensional PCA embedding, we observe a median absolute reconstruction error of 6.82 mm across the validation dataset.

One desirable feature of PCA is its consistency; the learning process is not susceptible to stochastic initialization and training procedures. The first 100 principal components are unchanging, regardless of whether we only want to use 50 to compress our data. We can efficiently explore the embedding dimension hyperparameter only running the PCA algorithm once. When median validation reconstruction errors are plotted against embedding dimension (Figure \ref{pcadimensionfig}), we notice an ``elbow'' indicating diminishing returns around 32-128 dimensions.

\subsection{Deep Autoencoders}

\begin{figure}[!hbt]
\centering
\begin{subfigure}{0.49\textwidth}
  \centering
  \includegraphics[width=\linewidth]{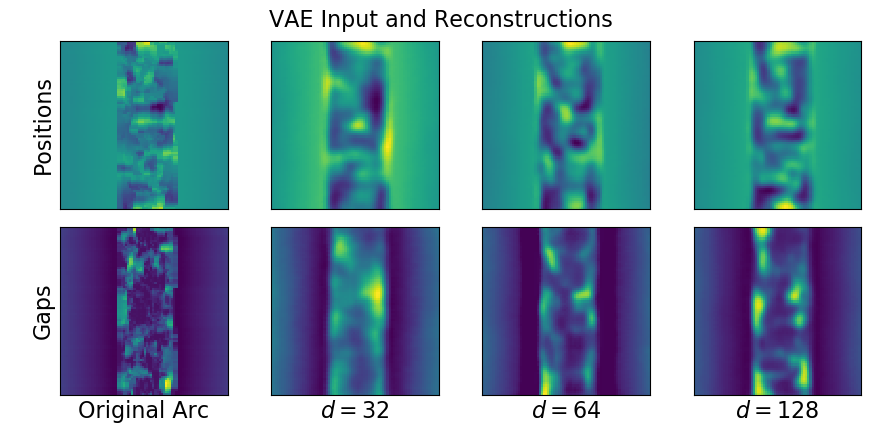}
\end{subfigure}
\begin{subfigure}{0.49\textwidth}
  \centering
  \includegraphics[width=\linewidth]{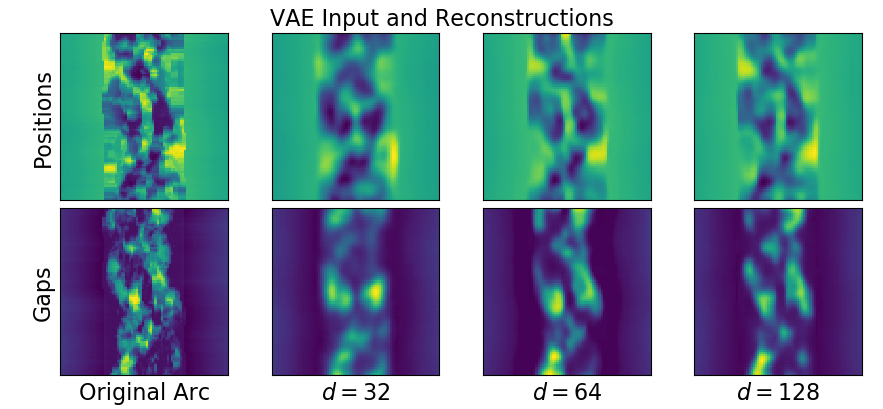}
\end{subfigure}
\caption[VAE Compression Results]{Validation arcs and their reconstructions with $k=32$ VAEs. Median absolute reconstruction error is reported in mm for each arc. Notice that higher embedding dimensions support increased modulation.}
\label{vaereconstructions}
\end{figure}

A family of 4-downblock models having $[k,\ 2k,\ 4k,\ 8k]$ filters per layer for hyperparameter $k$ was constructed for each autoencoder class (AE and VAE). Embedding dimensions $d \in \{32,\ 64,\ 128\}$ were considered. A detailed description of the models' architectures is given in Appendix \ref{archappendix}. Each model was trained using an Adam optimizer with a learning rate of 0.001 for a maximum of 500 epochs on the training dataset \cite{kingma2014adam}. The model state with the lowest 10-epoch running average validation loss was kept for deployment. Training examples were stochastically augmented online with random noise, horizontal, and vertical flipping. We consider dropout and dropblock regularization at 20\%. Following an initial comparison of these techniques with the $k=32$, $d=32$ model (respective median reconstruction errors of 1.46 and 1.40 mm), we use dropblock regularization for the entire model family. We also use batch normalization with a batch size of 64 \cite{ioffe2015batch}.  When training the variational autoencoders, we consider scaling the contribution of the KL divergence with hyperparameter $\alpha \in \{0.001,\ 0.01,\ 0.1,\ 1.0\}$ (Equation \ref{vaelossfn}). Figure \ref{vaereconstructions} shows images and their reconstructions for the $k=32$ VAEs having latent space dimensions 32, 64, and 128.

Notice that the models, especially those with lower embedding dimensions, forfeit some capacity to represent extreme regions of the planning space. Leaf positions which are far outside the norm are not represented in the learned latent space. Additionally, leaf positions are restricted to not vary drastically in either position or time; this is seen in the blurring of reconstructed arcs.

\begin{figure}[!hbt]\centering
    \begin {minipage}{0.49\textwidth}
     \centering
     \includegraphics[width=\linewidth]{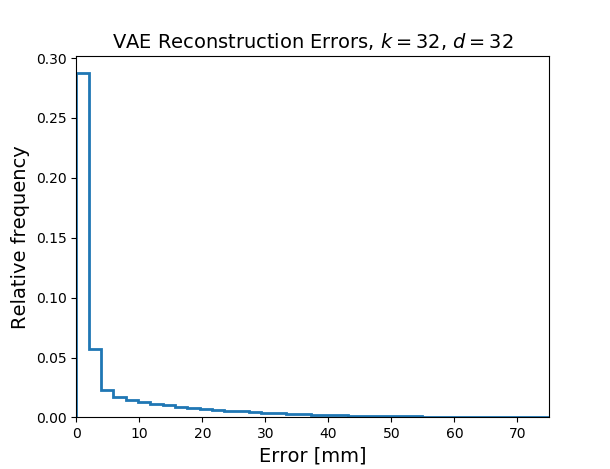}
     \caption[Compression Error Distribution]{Distribution of leaf position errors for reconstructed MLC arcs using the $k=32$, $d=32$ model. For this model, we find a median reconstruction error of 1.43 mm.}
     \label{reconerrors}
   \end{minipage}
   \hfill
   \begin{minipage}{0.49\textwidth}
     \centering
     \includegraphics[width=\linewidth]{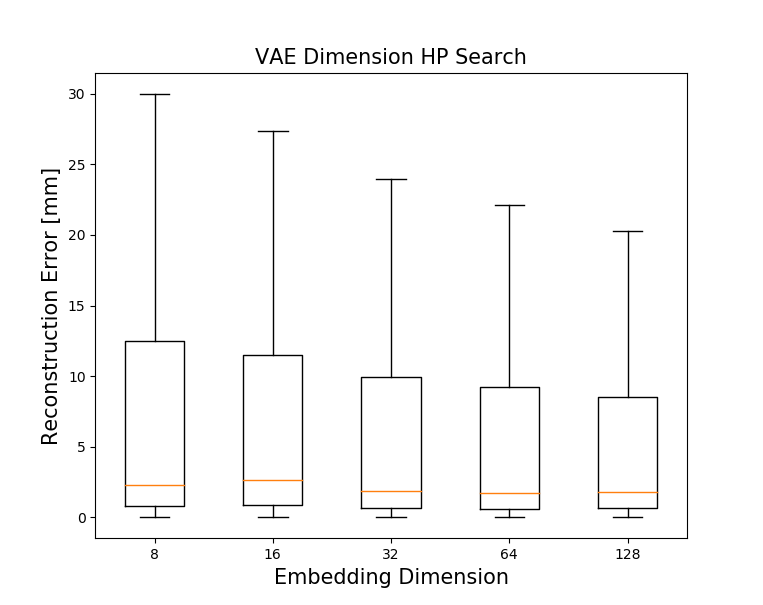}
     \caption[VAE Dimension Hyperparameter Search]{A trend towards more accurate reconstructions is observed for increasing latent space dimension. In this experiment, $k=32$ models were trained for 500 epochs each.}
     \label{zdimsearch}
   \end{minipage}
\end{figure}

The capacity to represent a greater variety of arcs is directly reflected in the latent space dimension $d$ (Figures \ref{pcadimensionfig} and \ref{zdimsearch}, Table \ref{compressiontable}). While representing more of the solution space promotes more accurate reconstructions and optimization solutions, it comes at the cost of an increased optimization search space. 

As in Figure \ref{pcaerrordist}, we can visualize the distribution of absolute errors for every pixel in the test dataset for a given model (Figure \ref{reconerrors}).  A median absolute error of 1.43 mm was observed with this model. Notice that the VAE error distribution contains significantly fewer errors in the 10-30 mm range than the distribution for PCA (Figure \ref{pcaerrordist}). 

Median absolute reconstruction errors are tabulated for each model in Table \ref{compressiontable}. For $d=64$ PCA, we observe a median absolute error of 6.82 mm compared to 1.12-1.76 mm for the $d=64$ VAEs. Linear machine learning largely fails to represent trends in the arc therapy dataset, at least in comparison to the deep learning models.

\begin{table}[!hbt]
\centering
\scalebox{1.05}{
\begin{tabular}{cccccccc}
\textbf{}                  & \textbf{}                  & \textbf{AE}                & \textbf{}                  & \textbf{}                  & \textbf{VAE}               &                            & \textbf{PCA}               \\
                           & $k=16$                       & $k=32$                       & $k=48$                       & $k=16$                       & $k=32$                       & $k=48$                       &   -                         \\ \cline{2-8} 
\multicolumn{1}{c|}{$d=32$}  & \multicolumn{1}{c|}{3.615} & \multicolumn{1}{c|}{4.559} & \multicolumn{1}{c|}{4.521} & \multicolumn{1}{c|}{2.596} & \multicolumn{1}{c|}{1.591} & \multicolumn{1}{c|}{1.513} & \multicolumn{1}{c|}{7.139} \\ \cline{2-8} 
\multicolumn{1}{c|}{$d=64$}  & \multicolumn{1}{c|}{4.292} & \multicolumn{1}{c|}{4.119} & \multicolumn{1}{c|}{3.827} & \multicolumn{1}{c|}{1.757} & \multicolumn{1}{c|}{1.430} & \multicolumn{1}{c|}{1.117} & \multicolumn{1}{c|}{6.064} \\ \cline{2-8} 
\multicolumn{1}{c|}{$d=128$} & \multicolumn{1}{c|}{5.394} & \multicolumn{1}{c|}{2.343} & \multicolumn{1}{c|}{3.742} & \multicolumn{1}{c|}{2.411} & \multicolumn{1}{c|}{1.585} & \multicolumn{1}{c|}{8.164} & \multicolumn{1}{c|}{5.068} \\ \cline{2-8} 
\end{tabular}}
\vspace{1pt}
\caption[Compression Results by Hyperparameter]{Autoencoders are able to compress and reproduce test set arc data with high accuracy compared to PCA. Median reconstruction error is reported in mm for both autoencoder models. Lower scores indicate a better performance.}\label{compressiontable}
\end{table}

%%%%%%%%%%%%%%%%%%%%%%%%%%%%%%%%%%%%%%%%%%%%%
\section{Building an Optimization Engine}\label{ch4}

The deep learning models are able to compress and reconstruct VMAT arcs with a high degree of accuracy compared to PCA. Every conventional autoencoder was outperformed by its variational counterpart in reconstructing validation arcs. For the compressions exhibited here, the VMAT search space exhibits a volume reduction by a factor of $N^{12800-d}$, where $N$ is the number of unique positions that each optimization parameter is allowed to occupy and $d$ is the embedding dimension. This is a noteworthy volume reduction and should be reflected in the total optimization time. In this section, we explore the capacity of the machine learning models to find solutions in two optimization settings.

First, we use a position-based objective function to test the models’ capacities to find meaningful arcs which are not explicitly represented in the training data. For this endeavor, we assume that held-out arcs from the validation dataset are optimal and evaluate the autoencoders' and PCA’s abilities find those solutions. Next, we test latent space optimization in a conventional radiotherapy planning setting with dose calculations and a dosimetric objective function. An esophageal cancer patient test case is used to demonstrate latent space optimization in progress.

\subsection{Latent Space Arc Therapy Optimization}

\begin{figure}[!hbt]
\centering
\makebox[\textwidth][c]{\includegraphics[width=0.8\linewidth]{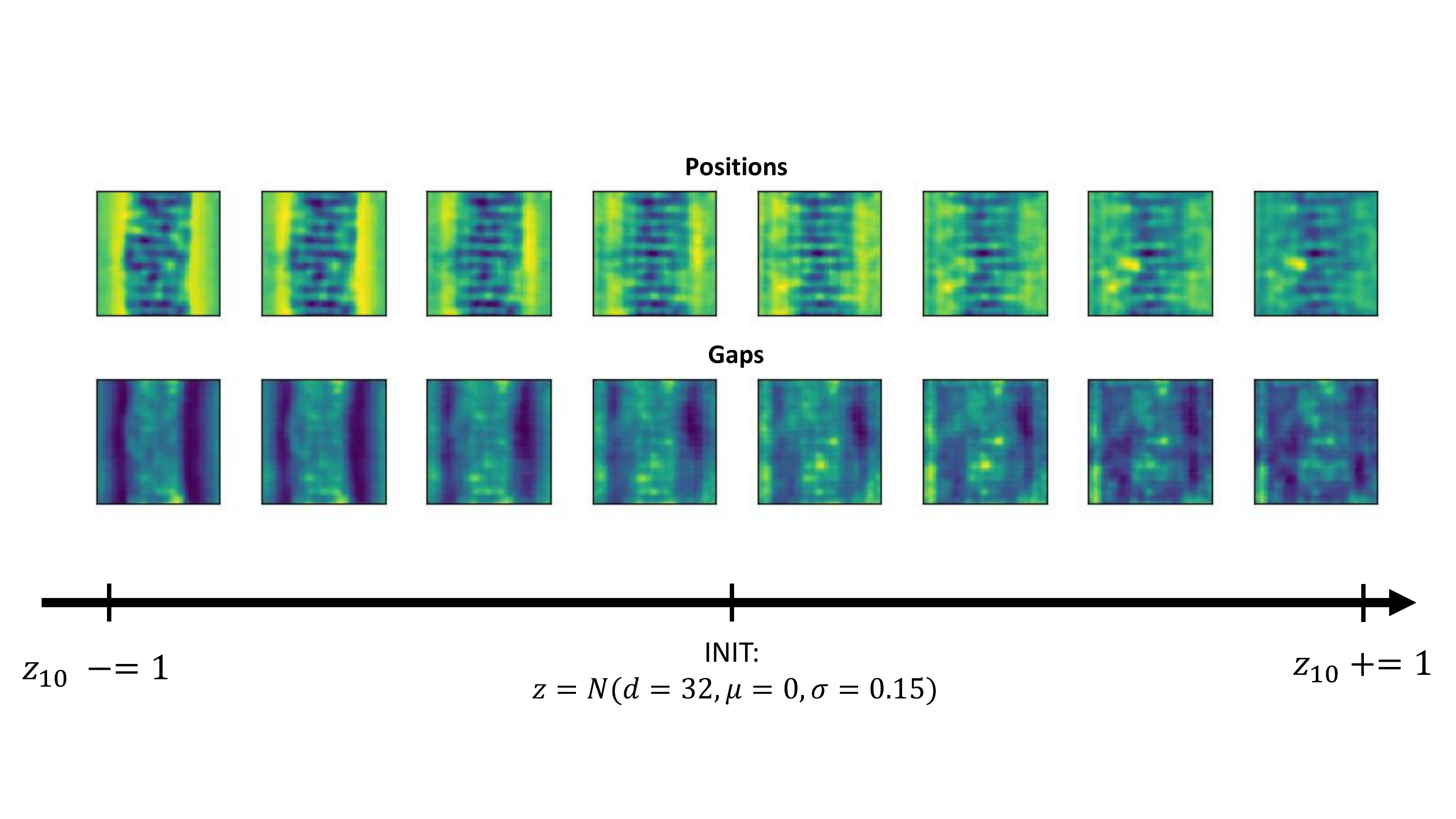}}
\caption[Fixing All But One Neuron]{Holding all but one neuron fixed, we see a smooth transition between arcs from $z_{10} \mathrel{-}= 1$ to $z_{10} \mathrel{+}= 1$. We can expect that there is some higher-level semantic meaning to this transition. An interactive widget demonstrating a similar process with the CelebA dataset and the Glow model is available at \url{https://openai.com/blog/glow/} \cite{kingma2018glow}.}
\label{varyonefig}
\end{figure}

In the previous section we tested the models' abilities to efficiently encode and decode training and validation arcs. The premise of latent space optimization, however, rests on two requirements. Training arcs should not only 1) be representable in the learned coordinates, but 2) have meaningful space between those coordinates. We would like to iteratively update some solution state by making small changes to it, and if modifying one coordinate in the latent space $\mathcal{Z}$ creates some large change in the input space $\mathcal{X}$, this might not be possible. We can show that small changes in $\mathcal{Z}$ correspond to small changes in $\mathcal{X}$ by fixing all but one latent space coordinate and visualizing variations in the decoder output. This is seen in Figure \ref{varyonefig}, in which we choose some point in $\mathcal{Z}$ near zero and consider small changes in the 10th direction (chosen arbitrarily). For this demonstration, we use the $d=32$, $k=32$ VAE. This is analogous to Figure 6 from Kingma \textit{et al.} and Figure 2c from Gomez-Bombarelli \textit{et al.} \cite{kingma2018glow, gomez2018automatic}.

\begin{figure}[!hbt]
  \centering
  \includegraphics[width=0.65\linewidth]{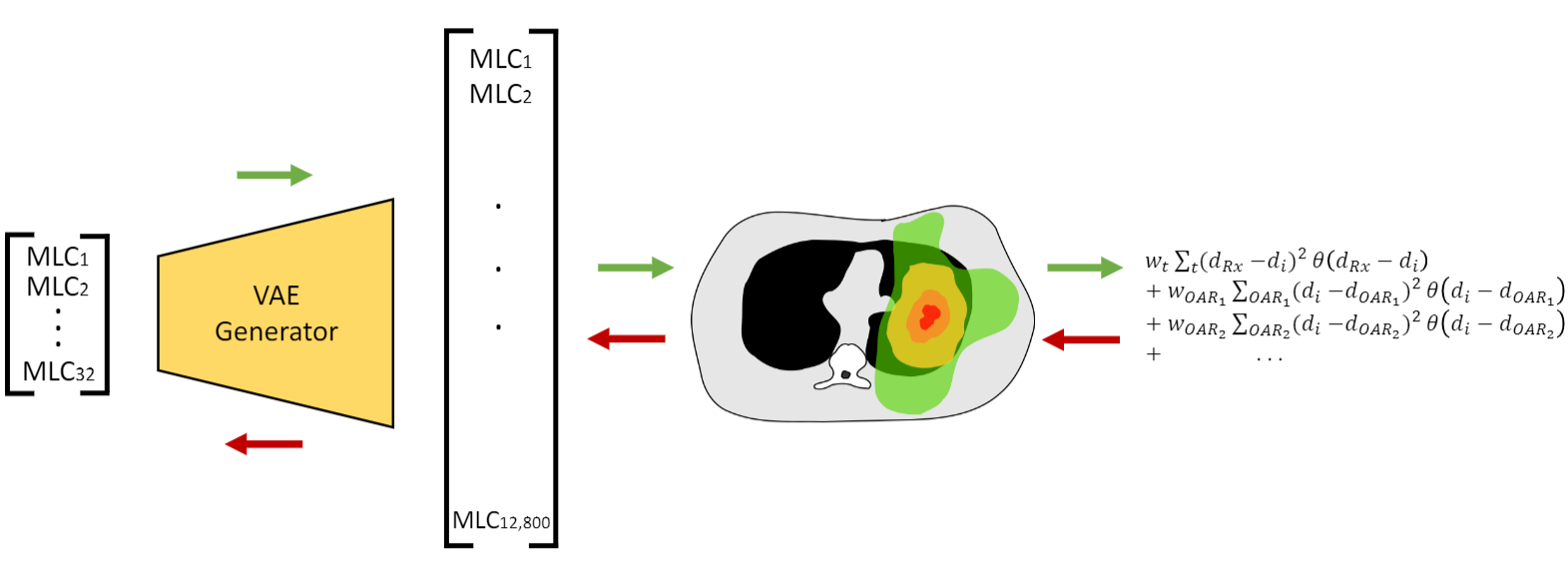}
\caption[VMAT Optimization Framework]{The proposed VMAT optimization framework is visualized. In conventional arc therapy optimization, MLC leaf positions are used to calculate dose distributions from which dosimetric objectives are evaluated. Error computations are used to inform the modification of the arc according to some algorithm such as simulated annealing or gradient descent. In latent space optimization, an autoencoder generator is introduced to translate between the low- and high-dimensional arc parameterizations.}
\label{optimizerfigure}
\end{figure}

The proposed arc therapy optimization framework is visualized in Figure \ref{optimizerfigure}. This routine is identical to conventional VMAT optimization routines except for the coordinate transformation provided by the decoding portion of an autoencoder. In the graphic, green arrows are used to represent the forward-propagation of an error calculation given coordinates $z$, while red arrows represent the propagation of error backwards through the system in a gradient calculation. In the coming experiments, stochastic optimization is used exclusively. Gradient-based implementations of latent space optimizers are discussed in Appendix \ref{gradsection}.

To remove confounding variables from our optimization experiments, we perform them in a controlled setting by removing the dose calculations. For these experiments, we test the various optimizers' capacities to find held-out validation arcs according to a leaf-position-based optimization objective (Section \ref{posexperiments}).

When VMAT algorithms are compared, remember that the selection of an algorithm for clinical use is not entirely objective. The problem of choosing an optimization algorithm can itself be formulated as an optimization problem. We can imagine some physician satisfaction penalty function $O$ which depends on the optimization time $\tau$ and IMRT objective $f$. For example, $f$ might be prioritized for stereotactic treatments while $\tau$ is prioritized for daily adaptive therapy. Then, for our family of VAE's having latent space dimensions $d$, we might perform some algorithm selection based on $O$. While algorithms which accommodate low $f$ are currently clinically available, this work aims to provide some algorithms which permit low $\tau$ in addition to low $f$. Additionally, we aim to provide a mechanism of implicitly choosing which objective is prioritized (accuracy or time) with the selection of an optimization algorithm specified by its latent space dimension $d$.

\subsection{Position-based Optimization Experiments}\label{posexperiments}

Each VMAT optimization algorithm employed in this work uses simulated annealing to search its solution space. Three solution spaces are considered: 1) the fully parameterized arc space used in global DAO, 2) the latent space learned with deep autoencoders, and 3) the latent space learned with PCA. No gradient information is required during the implementation of simulated annealing.

Each algorithm is used to find VMAT arcs according to a position-based objective function defined by the square displacement of each leaf from its target position. For state $s$ and target $s^*$, the error is defined\footnote{Recall that each plan is represented with $2*80*80=12,800$ leaf positions in our parameterization.}
\begin{equation}\label{msedisplacement}
    E(s) = \frac{1}{12,800}\sum_{i = 1}^{12,800} (s_i - s^*_i)^2 \ .
\end{equation}
The algorithms are given the same challenge of finding arcs from the held-out validation dataset (188 examples). We allow the optimizers to search with a decreasing temperature until the error decreases by less than 1\% over 15 seconds. When this condition is met, the algorithms are halted and their performance is reported. 

\subsubsection{Measuring Optimizer Performance}
Following an optimization experiment, we can identify the absolute position error for every leaf according to
\begin{equation}
    \epsilon_i = |s_i - s^*_i| \ .
\end{equation}
We then report the optimizer's performance as the median absolute position accuracy $\Tilde{\epsilon}$ and its standard error $\sigma_{\Tilde{\epsilon}}$ over the validation dataset. Additionally, we track and visualize the reduction of the normalized MSE displacement (Equation \ref{msedisplacement}) during every trial.

The greatest expected benefit for using a low-dimensional representation of arcs during optimization is a reduction in the number of iterations required for the algorithm to converge. For every optimization, we monitor the total number of iterations $N$ required before the $<$ 1\% change in error criterion is met. The median number of iterations $\Tilde{N}$ and its standard error $\sigma_{\Tilde{N}}$ are monitored.

We also report the median total optimization time $\Tilde{\tau}$ which is required by each optimizer and standard error $\sigma_{\Tilde{\tau}}$. We recognize that the total optimization time in these experiments might not entirely reflect optimization times in a clinical setting. For any VMAT optimization algorithm, we expect that the total optimization time before convergence $\tau$ can be represented as 
\begin{equation}
    \tau = N(T_O + T_D)
\end{equation}
where $T_D$ is the time spent at every iteration on dose calculations and $T_O$ is the time spent on all other optimizer calculations. For the latent space optimizers, there is significant computational overhead of having to run the AE generator at every iteration. During the experiments with a leaf-position-based objective where $T_D=0$, the total optimization time depends equally on $N$ and $T_O$. However, in a clinical setting where $T_D$ can be large, there is a trade-off between prioritizing low $N$ or low $T_O$. To demonstrate the effect of $T_D$ on the total optimization time we perform additional experiments simulating dose calculations by suspending executions at every iteration for some $T_D$. For $T_D$ simulation, we consider values from 10-100 ms.

\subsubsection{Position-based Optimization Results}
\begin{figure}[!hbt]
\centering
\begin{subfigure}{0.32\textwidth}
  \centering
  \includegraphics[width=\linewidth]{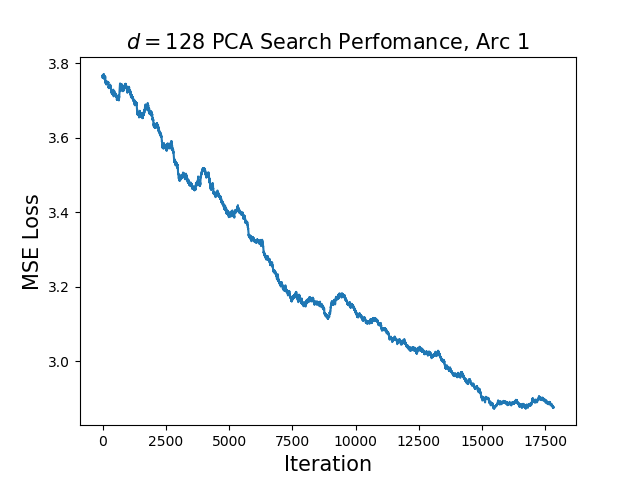}
\end{subfigure}
\begin{subfigure}{0.32\textwidth}
  \centering
  \includegraphics[width=\linewidth]{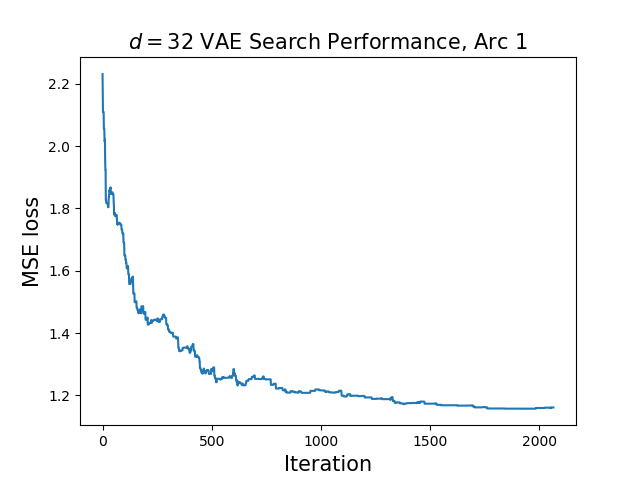}
\end{subfigure}
\centering
\begin{subfigure}{0.32\textwidth}
  \centering
  \includegraphics[width=\linewidth]{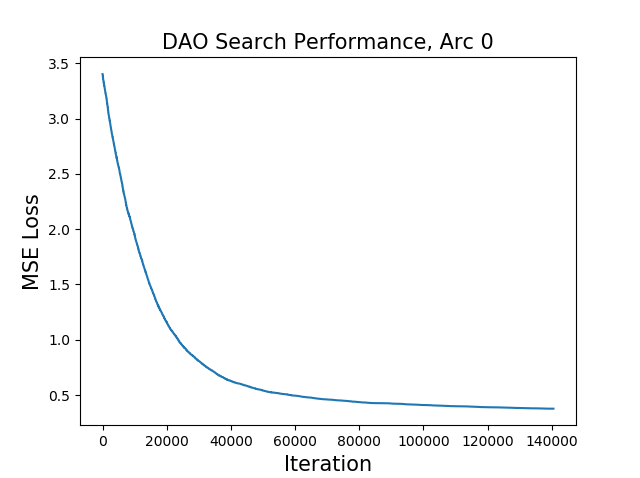}
\end{subfigure}
\caption[Reductions in Optimization Objective]{The MSE objective is reduced over the course of optimization with simulated annealing in each search space. Here, error reductions are shown for each algorithm in finding the first held-out arc from the validation dataset.}
\label{reductioninobjective}
\end{figure}

For every algorithm, we can visualize the reduction in MSE over the course of optimization for a given arc (Figure \ref{reductioninobjective}). Each optimizer is allowed to search for held-out arcs until convergence, and metrics describing the optimization are collected. Table \ref{results} displays the median error $\Tilde{\epsilon}$, the median number of iterations $\Tilde{N}$, and the median optimization time $\Tilde{\tau}$ using the 32-128 dimensional latent spaces learned with the AE, VAE, and PCA. The performance of global DAO with mock geometry-informed initialization is also shown for comparison.

We find that PCA is not able to find solutions with satisfactory errors, having $\Tilde{\epsilon} \approx$ 6 mm even for the 128 dimensional latent space. In general, the VAEs perform considerably better than their non-variational counterparts, with the models for every latent space dimension converging with lower error in fewer iterations. We find that of all the models, the $d=32$ VAE latent space algorithm converges in the fewest iterations ($\Tilde{N} = 2,766 \pm 63$), while the $d=128$ VAE achieves the lowest median absolute error before converging ($\Tilde{\epsilon} = 1.583 \pm 0.753$ mm). $d=32$ PCA converges the fastest, but does so to poor quality solutions.

\begin{table}
\centering
\resizebox{0.7\textwidth}{!}{
\begin{tabular}{ccccccc}
                                                                                                                 & \multicolumn{1}{c}{$\Tilde{\epsilon}$ {[}mm{]}} & \multicolumn{1}{c}{$\sigma_{\Tilde{\epsilon}}$ {[}mm{]}} & $\Tilde{N} [10^3]$                        & $\sigma_{\Tilde{N}} [10^3]$         & $\Tilde{\tau}$ {[}s{]}             & $\sigma_{\Tilde{\tau}}$ {[}s{]} \\ \cline{2-7} 
\multicolumn{1}{c|}{\textbf{AE $\mathbf{d=32}$}}                                                                          & \multicolumn{1}{c|}{2.844}                              & \multicolumn{1}{c|}{1.002}                             & \multicolumn{1}{c|}{3.138}         & \multicolumn{1}{c|}{0.589} & \multicolumn{1}{c|}{69.6}          & \multicolumn{1}{c|}{1.15}    \\ \cline{2-7} 
\multicolumn{1}{c|}{\textbf{AE $\mathbf{d=64}$}}                                                                          & \multicolumn{1}{c|}{2.733}                              & \multicolumn{1}{c|}{0.831}                             & \multicolumn{1}{c|}{4.728}         & \multicolumn{1}{c|}{0.081} & \multicolumn{1}{c|}{86.34}          & \multicolumn{1}{c|}{1.49}    \\ \cline{2-7} 
\multicolumn{1}{c|}{\textbf{AE $\mathbf{d=128}$}}                                                                         & \multicolumn{1}{c|}{1.960}                              & \multicolumn{1}{c|}{0.695}                             & \multicolumn{1}{c|}{7.157}         & \multicolumn{1}{c|}{0.105} & \multicolumn{1}{c|}{143.1}          & \multicolumn{1}{c|}{2.15}    \\ \cline{2-7} 
\multicolumn{1}{c|}{\textbf{VAE $\mathbf{d=32}$}}                                                                         & \multicolumn{1}{c|}{1.959}                              & \multicolumn{1}{c|}{0.921}                             & \multicolumn{1}{c|}{\textbf{2.766}} & \multicolumn{1}{c|}{0.063} & \multicolumn{1}{c|}{50.42}          & \multicolumn{1}{c|}{1.08}    \\ \cline{2-7} 
\multicolumn{1}{c|}{\textbf{VAE $\mathbf{d=64}$}}                                                                         & \multicolumn{1}{c|}{1.872}                     & \multicolumn{1}{c|}{0.890}                             & \multicolumn{1}{c|}{3.306}          & \multicolumn{1}{c|}{0.078} & \multicolumn{1}{c|}{79.3}          & \multicolumn{1}{c|}{1.87}    \\ \cline{2-7} 
\multicolumn{1}{c|}{\textbf{VAE $\mathbf{d=128}$}}                                                                        & \multicolumn{1}{c|}{\textbf{1.583}}                              & \multicolumn{1}{c|}{0.753}                             & \multicolumn{1}{c|}{4.769}          & \multicolumn{1}{c|}{0.092} & \multicolumn{1}{c|}{103.4}          & \multicolumn{1}{c|}{2.03}    \\ \cline{2-7} 
\multicolumn{1}{c|}{\textbf{PCA $\mathbf{d=32}$}}                                                                         & \multicolumn{1}{c|}{6.900}                              & \multicolumn{1}{c|}{1.047}                             & \multicolumn{1}{c|}{89.23}         & \multicolumn{1}{c|}{0.236}    & \multicolumn{1}{c|}{\textbf{22.5}} & \multicolumn{1}{c|}{0.08} \\ \cline{2-7} 
\multicolumn{1}{c|}{\textbf{PCA $\mathbf{d=64}$}}                                                                         & \multicolumn{1}{c|}{7.082}                              & \multicolumn{1}{c|}{1.137}                             & \multicolumn{1}{c|}{54.14}         & \multicolumn{1}{c|}{0.397}    & \multicolumn{1}{c|}{39.77}          & \multicolumn{1}{c|}{0.75} \\ \cline{2-7} 
\multicolumn{1}{c|}{\textbf{PCA $\mathbf{d=128}$}}                                                                        & \multicolumn{1}{c|}{7.082}                              & \multicolumn{1}{c|}{1.137}                             & \multicolumn{1}{c|}{53.62}         & \multicolumn{1}{c|}{0.334}    & \multicolumn{1}{c|}{36.65}           & \multicolumn{1}{c|}{0.46} \\ \cline{2-7} 
\multicolumn{1}{c|}{\textbf{\begin{tabular}[c]{@{}c@{}}Global DAO with \\ Geometry Initialization\end{tabular}}} & \multicolumn{1}{c|}{5.593}                              & \multicolumn{1}{c|}{0.445}                             & \multicolumn{1}{c|}{127.23}         & \multicolumn{1}{c|}{1.973}    & \multicolumn{1}{c|}{88.6}          & \multicolumn{1}{c|}{6.09} \\ \cline{2-7} 
\end{tabular}}
\vspace{2pt}
\caption[Position-based Optimization Results]{All optimization routines are compared on the held-out validation dataset (188 examples). We find that the autoencoder optimizers, particularly those using a VAE generator, achieve lower error in fewer iterations compared to conventional global DAO. PCA, on the other hand, appears to be unable to find satisfactory solutions in a reasonable number of iterations, likely due to a limited expression of arcs.}\label{results}
\end{table}

\begin{figure}[!hbt]
  \centering
  \makebox[\textwidth][c]{\includegraphics[width=0.8\linewidth]{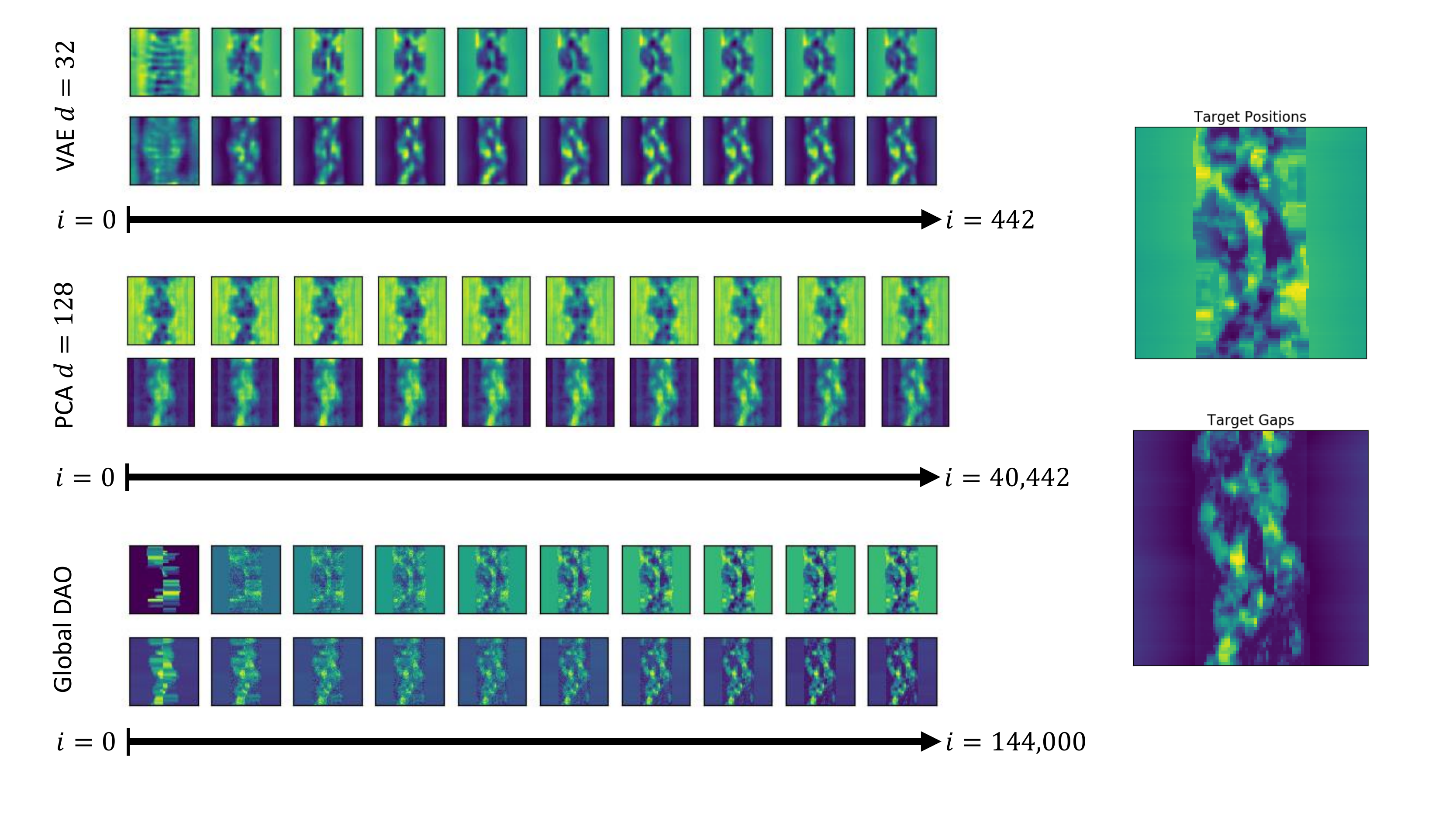}}
\caption[Visualizing the Progression of Algorithm States]{The state of each optimizer as it develops visualized over the course 10 evenly-spaced frames from iteration $i=0$ onward until convergence. Random steps in the global DAO parameterization create fuzzy images over the course of many iterations, while the VAE latent space optimizer takes meaningful steps and converges quickly.}
\label{smear}
\end{figure}

The search process can be visualized by displaying the arcs' states as images over the course of optimization. In Figure \ref{smear}, the process of each VMAT optimizer is visualized for the same target held-out arc from the validation set. While small random steps in the fully-parameterized planning space are not guaranteed to be clinically reasonable, the autoencoder limits the search space such that only reasonable plans are considered. This is seen in the fuzzy appearance of the developing DAO solution compared to the VAE solution.

\begin{figure}[!hbt]
  \centering
  \includegraphics[width=0.6\linewidth]{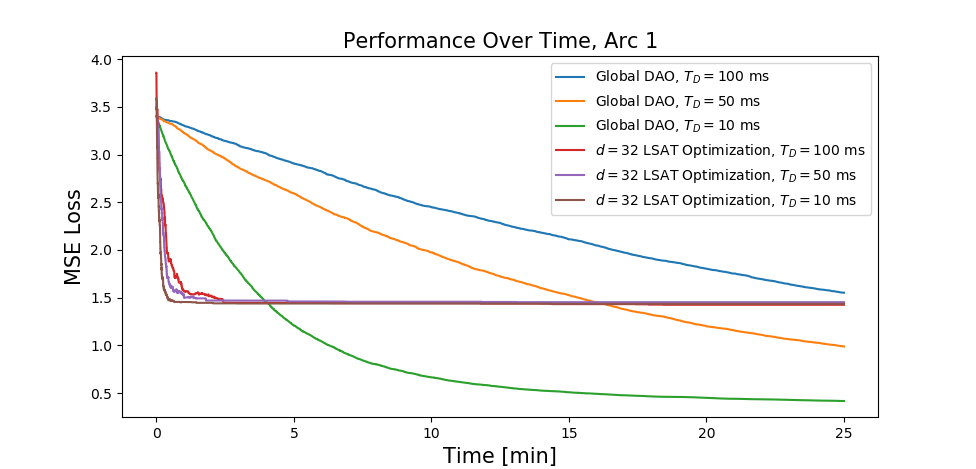}
\caption[Position-based Optimizer Comparison]{The MSE objective is reduced over the course of optimization for various simulated dose calculation times $T_D$. Despite the latent space optimizers' rapid convergence, the fully-parameterized DAO can achieve lower objective scores given enough time.}
\label{positionerrortime}
\end{figure}

Finally, we compare the performance of the $d=32$ VAE and global DAO algorithms with simulated dose calculation times $T_D$ for an arc from the validation dataset (randomly selected). In each case, the algorithms are allotted 25 minutes to reduce the MSE objective (for the normalized data) for a given target arc. Results are shown in Figure \ref{positionerrortime}. In general, we find that the latent space optimizer converges much faster. However, given enough time, global DAO has a higher capacity for modeling the ``corners'' of the planning space and can achieve lower objective scores. This observation might inform the future implementation of some logic where a latent space optimizer is used for the first few iterations followed by global DAO or local gradient-based DAO.

\subsection{Dose-based Optimization}

Latent space optimization was implemented with a $d=64$ VAE in the conventional clinical setting with dosimetric objective functions as illustrated in Figure \ref{optimizerfigure}. For comparison with a traditional optimizer, global DAO with geometry-based segment initialization was also implemented. An esophageal cancer patient's image data and structures were gathered from the Head-and-Neck-Radiomics-HN1 dataset from the Cancer Imaging Archive \cite{wee2019data}. 3D dose calculations were performed with a pencil beam algorithm to construct dose influence matrices. Dose was then calculated according to Equation \ref{dosesum}. Typical head and neck dose constraints were prescribed as recommended by an experienced physician and optimization was performed. Errors for each voxel were specified by their square deviation from their target values.

An esophageal cancer patient's image data and structures were gathered from the Head-and-Neck-Radiomics-HN1 dataset from the Cancer Imaging Archive \cite{wee2019data}. A dose influence matrix for a 10 cm x 10 cm x 10 cm dose grid in 3.3 mm x 3.3 mm x 3.3 mm resolution was calculated about the target's centroid (coincident with the machine's geometric isocenter). During optimization, beam weights were stochastically updated at every iteration along with the solution state via simulated annealing. At the beginning and end of the stochastic search, beam weights were refined with gradient-based beam weight optimization. 

\begin{figure}[htb]
\centering
% \begin{subfigure}{0.49\textwidth}
%   \centering
%   \includegraphics[width=0.8\linewidth]{figures/dao_error.png}
% \end{subfigure}
% \begin{subfigure}{0.49\textwidth}
%   \centering
%   \includegraphics[width=0.8\linewidth]{figures/vae_error.png}
% \end{subfigure}
  \centering
  \includegraphics[width=0.65\linewidth]{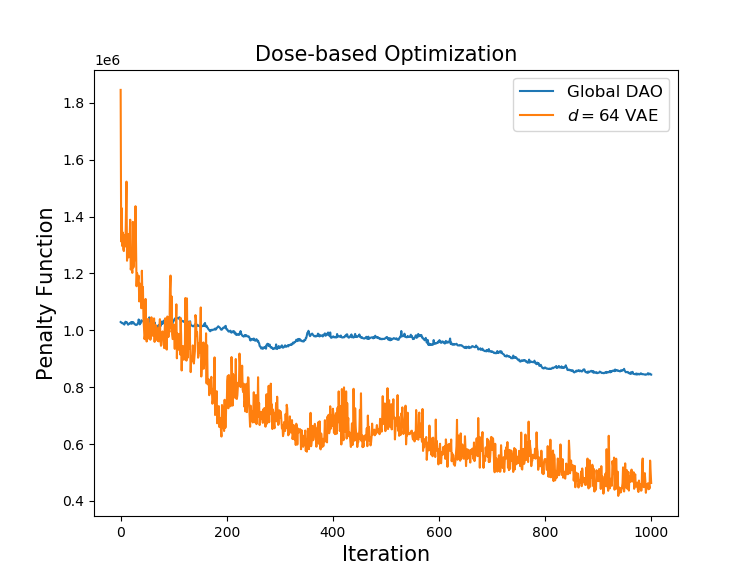}
\caption[Dose-based Optimization]{The dosimetric penalty function is reduced over the course of the search with each optimization algorithm in a clinical setting for an esophageal treatment.}
\label{doseoptresults}
\end{figure}

The global DAO and $d=64$ VAE latent space algorithms were allowed to run for 1,000 iterations. Over the course of optimization, both algorithms were able to significantly reduce the dosimetric penalty function (Figure \ref{doseoptresults}). Given the same number of iterations, latent space optimization is able to reduce the penalty function by 75\%, compared to global DAO's reduction of 8\%. 

\begin{figure}[!htb]
\centering
\makebox[\textwidth][c]{\includegraphics[width=0.65\linewidth]{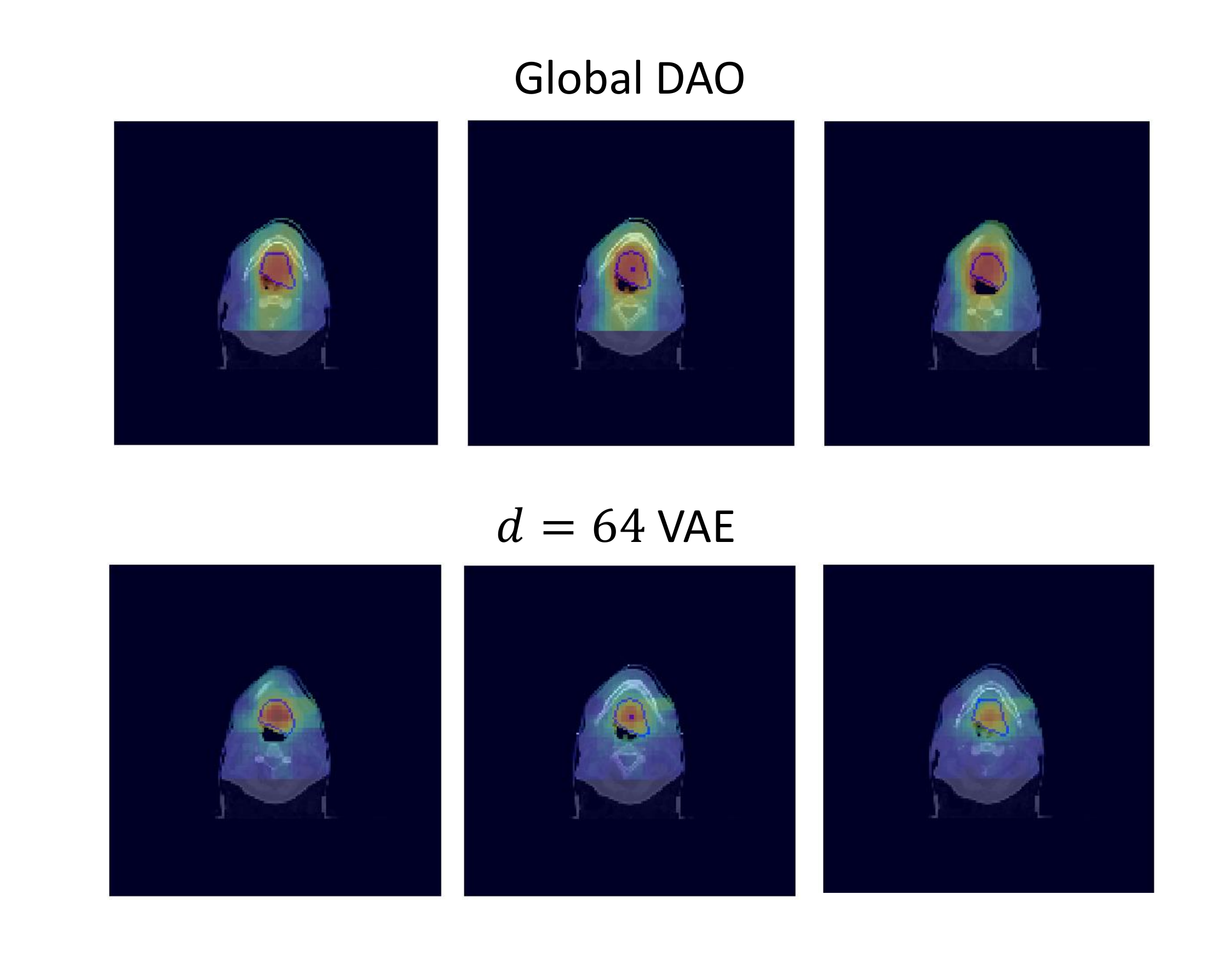}}
\caption[DVH Change]{Dose distributions for plans created with each optimizer. The target in these images is outlined in blue.}
\label{dosedist}
\end{figure}

While neither plan is clinically acceptable after 1,000 iterations, the potential of using a reduced search space is fairly demonstrated here, and we expect that either algorithm might be improved with fine-tuning via gradient-based local DAO. The dose distributions achieved by each algorithm are shown over the length of the target in Figure \ref{dosedist}.

\section{Discussion}
We found that, compared to conventional global DAO with geometry-based segment initialization, latent space optimization with VAEs significantly reduces the number of iterations required to find satisfactory arc therapy plans. This, in turn, reduces the total required optimization time. In the position-based optimization experiments with a 1\% change in error stopping criterion, the 128-dimensional VAE finds the highest accuracy solution of all of the algorithms with $\Tilde{\epsilon} = 1.583$, while 32-dimensional VAE converges in the fewest iterations $\Tilde{N} = 2766$ (Table \ref{results}). Given enough time, global stochastic DAO was able to achieve higher accuracy solutions than the latent space optimizers, but the VAE-based algorithms found the neighborhood of a solution much faster. 

Given 1,000 iterations to reduce a dosimetric penalty function, we find that the latent space search appears to be much more efficient, with error reductions of about 75\% compared to 18\% for global DAO. While neither plan is clinically acceptable after 1,000 iterations, the VAE-assisted algorithm more significantly reduces the error during this time.

While the conventional, non-variational AEs are able to sufficiently represent held-out arcs, they are generally outperformed by the variational AEs. Similarly, held-out arcs are representable to some degree with linear PCA, but not well enough for satisfactory VMAT optimization. It is likely that other machine learning models, such as generative adversarial networks, might be better-suited to learning the arc subspace than even VAEs.

A novelty of this work is the provision of a choice of what is clinically desired from the optimizer. Rapid daily adaptive planning might be possible with a 10-dimensional latent space, while more complex treatments might require 100+ dimensions. At the least, current clinical planning systems might benefit from some initializations informed with latent space optimization -- gradient based or stochastic. For this, we recommend the use of a low-dimensional representation until the optimizer begins to see diminished returns, at which point a search of the full space can be implemented.

A direction which we did not explore here is the comparison of a random initialization versus geometry-based initialization for the latent space algorithms. In our experiments, a random point near zero in the latent space $\mathcal{Z}$ was used to initialize the search. We expect that some geometry-based segment initialization can be represented in $\mathcal{Z}$ with the use of the encoding portion of the AEs. This might improve solution quality, especially for higher latent space dimensions.

One indirect consequence of searching for arcs in low-dimensional spaces is that higher resolution dose calculations are permitted. As we see in Figure \ref{positionerrortime}, the VAE algorithms are less affected by larger dose calculation times due to their ability to converge in relatively few iterations. This might permit higher beamlet and dose grid resolutions in future optimization settings.

Finally, we expect that the concept of pattern recognition for machine parameters might be applicable to other modalities. Other large solution spaces such as those in $4\pi$-radiotherapy might be reduced with deep-learning-based dimensionality reduction.

%%%%%%%%%%%%%%%%%%%%%%%%%%%%%%%%%%%%%%%%%%%%%%%%%%%%%%%%
\newpage
\begin{appendices}
\numberwithin{table}{section}
\numberwithin{figure}{section}
\section{Autoencoder Architecture}\label{archappendix}

\begin{table}[!hbt]
\centering
\resizebox{\textwidth}{!}{
\begin{tabular}{|c|c|c|c|c|}
\hline
\textbf{Block}            & \textbf{Sub-block} & \textbf{Layers}                    & \textbf{Output Shape} & \textbf{Kernels}                     \\ \hline
Input                     & -                  & -                                  & (2, 80, 80)           & -                                    \\ \hline
\multirow{4}{*}{Encoder}  & Downblock1         & Conv2D, MaxPool2D, DB, BN, ReLU    & (k, 40, 40)           & $k=(3\mbox{x}3), s=1,  k_p=(2\mbox{x}2), s_p=2, D=0.2$ \\ \cline{2-5} 
                          & Downblock2         & Conv2D, MaxPool2D, DB, BN, ReLU    & (2k, 20, 20)          & $k=(3\mbox{x}3), s=1,  k_p=(2\mbox{x}2), s_p=2, D=0.2$ \\ \cline{2-5} 
                          & Downblock3         & Conv2D, MaxPool2D, DB, BN, ReLU    & (4k, 10, 10)          & $k=(3\mbox{x}3), s=1,  k_p=(2\mbox{x}2), s_p=2, D=0.2$ \\ \cline{2-5} 
                          & Downblock4         & Conv2D, MaxPool2D, DB, BN, ReLU    & (8k, 5, 5)            & $k=(3\mbox{x}3), s=1,  k_p=(2\mbox{x}2), s_p=2, D=0.2$  \\ \hline
\multirow{4}{*}{Sampling} & Flattening         & Flatten                            & (8k * 5 * 5)          & -                                    \\ \cline{2-5} 
                          & To means / vars    & Linear, ReLU, Reshape              & (2, d)                & -                                    \\ \cline{2-5} 
                          & To z               & Sample                             & (d)                   & -                                    \\ \cline{2-5} 
                          & To image           & Linear, ReLU, Reshape              & (8k, 5, 5)            & -                                    \\ \hline
\multirow{4}{*}{Decoder}  & Upblock1           & ConvTranspose2D, DB, BN, ReLU      & (4k, 10, 10)          & $k=(3\mbox{x}3), s=2, D=0.2$                  \\ \cline{2-5} 
                          & Upblock2           & ConvTranspose2D, DB, BN, ReLU      & (2k, 20, 20)          & $k=(3\mbox{x}3), s=2, D=0.2$                  \\ \cline{2-5} 
                          & Upblock3           & ConvTranspose2D, DB, BN, ReLU      & (k, 40, 40)           & $k=(3\mbox{x}3), s=2, D=0.2$                  \\ \cline{2-5} 
                          & Upblock4           & ConvTranspose2D, DB, BN            & (2, 80, 80)           & $k=(3\mbox{x}3), s=2, D=0.2$                  \\ \hline
Output                    & -                  & Linear (channel 1), ReLU (channel 2) & (2, 80, 80)           & -                                    \\ \hline
\end{tabular}}
\vspace{2pt}
\caption[VAE Architecture]{An example VAE architecture for hyperparameters $k$, $d$, and $D$. For use in the optimizer, we use the generator specified by ``To image'' and all subsequent layers. Here, DB and BN are used to represent dropblock and batch normalization layers, respectively. $k$ and $s$ are used to designate convolution layer kernel size and stride, while $k_p$ and $s_p$ are used to designate pooling layer kernel size and stride. In each dropblock layer, dropout rate $D$ is used.}\label{vaearchitecture}
\end{table}

\section{Mock Geometry-based Segment Initialization}
Our implementation of global DAO in the position-based optimization setting is intended to replicate the procedure outlined by Earl \textit{et al.} \cite{earl2003inverse}. Because no patient-specific image information is provided for each arc, the global DAO algorithm's state is initialized with leaf positions based on the solution leaf positions (or gaps) within each aperture at every control point. The detailed initialization procedure is shown in Algorithm \ref{geoinit}.
% \vspace{0.5 cm}
\begin{figure}[H]
  \centering
  \begin{minipage}{.8\linewidth}
    \begin{center}
    \begin{algorithm}[H]
    \SetAlgoLined
    Initialize state $s$, solution $s^*$, $\dim(s) = \dim(s^*) = (80, 80)$\\
    \For{$\theta \gets 0$ \KwTo $80$}{
            \For{$l \gets 0$ \KwTo $40$}{
                $s[\theta, l+1] \gets \max(s[\theta, l], s^*[\theta, l+1])$\\
                $s[\theta, -(l+1)] \gets \max(s[\theta, -l], s^*[\theta, -(l+1)])$
                }
            }
    \Return s
    \caption[Mock Geometry-informed Initialization]{Mock Geometry-informed Initialization}\label{geoinit}
    \end{algorithm}
    \end{center}
    \end{minipage}
    \end{figure}
% \vspace{0.5 cm}

\begin{figure}[!hbt]
\centering
\begin{subfigure}{0.24\textwidth}
  \centering
  \includegraphics[width=\linewidth]{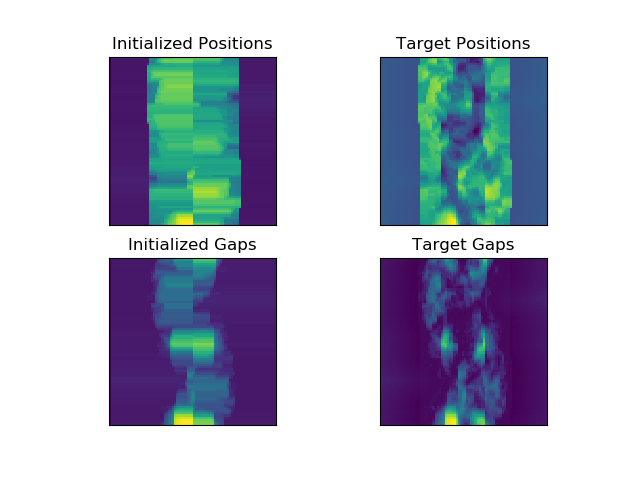}
\end{subfigure}
\begin{subfigure}{0.24\textwidth}
  \centering
  \includegraphics[width=\linewidth]{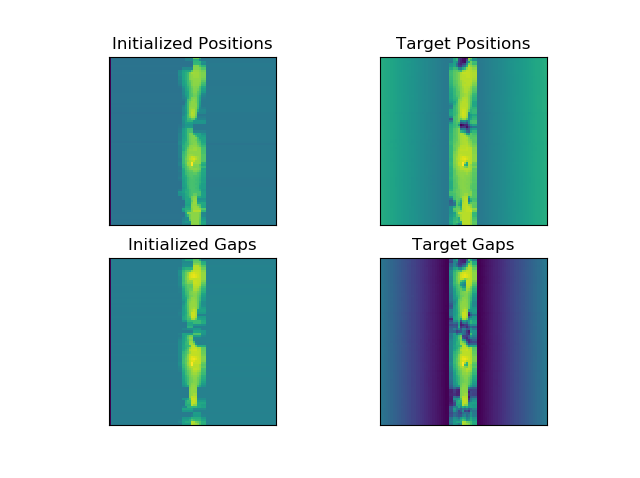}
\end{subfigure}
\begin{subfigure}{0.24\textwidth}
  \centering
  \includegraphics[width=\linewidth]{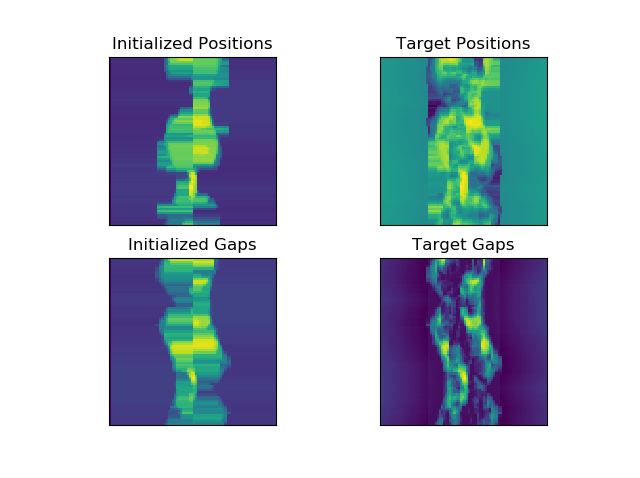}
\end{subfigure}
\begin{subfigure}{0.24\textwidth}
  \centering
  \includegraphics[width=\linewidth]{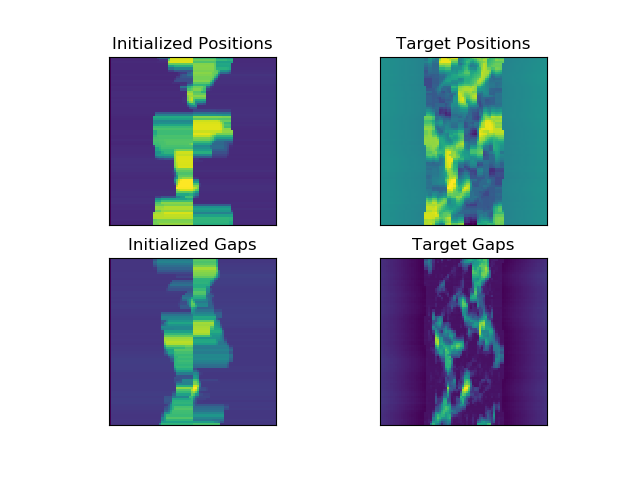}
\end{subfigure}
\caption[Geometry Initialization Approximations]{Some ``geometry-based'' arc initializations. For each target arc from the validation dataset, a geometry-based initialization is shown. These serve as a starting point for the search of the full VMAT solution space using global stochastic DAO.}
\label{geometryinit}
\end{figure}

\begin{figure}[!hbt]
\centering
\begin{subfigure}{0.32\textwidth}
  \centering
  \includegraphics[width=\linewidth]{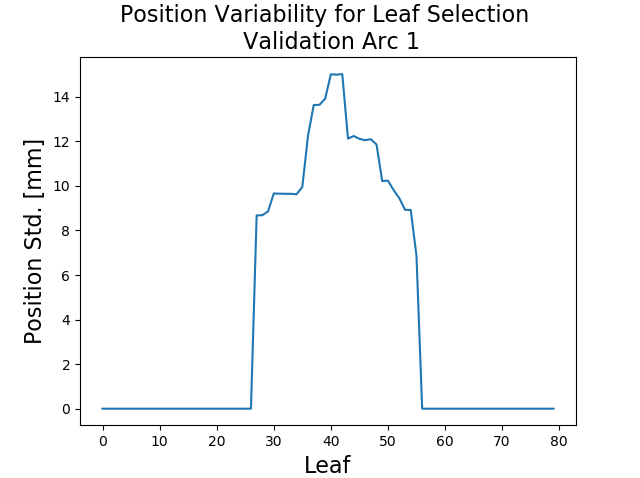}
\end{subfigure}
\begin{subfigure}{0.32\textwidth}
  \centering
  \includegraphics[width=\linewidth]{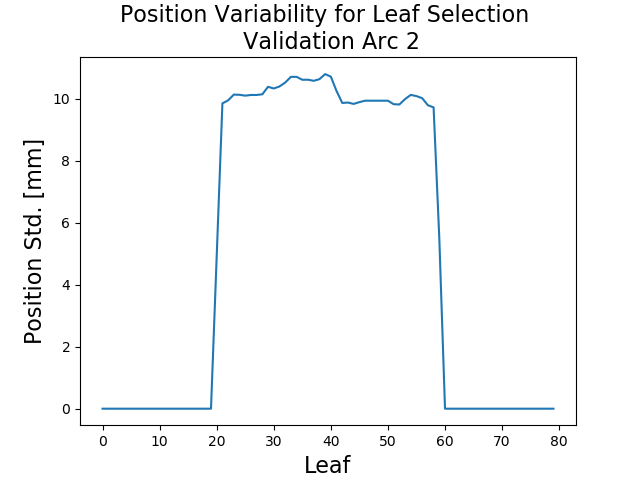}
\end{subfigure}
\centering
\begin{subfigure}{0.32\textwidth}
  \centering
  \includegraphics[width=\linewidth]{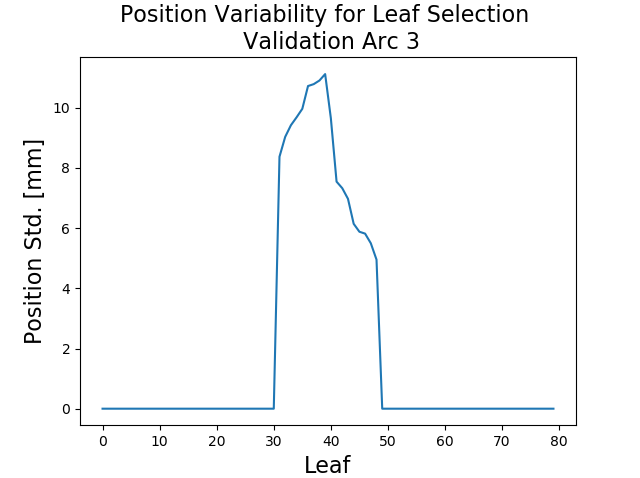}
\end{subfigure}
\caption[Leaf Selection for Global DAO]{Spread of leaf positions over the duration of the arc after geometry initialization. A region of leaves is selected for global DAO based on these distributions.}\label{leafvariability}
\end{figure}

\noindent This provides a fair approximation to the solution arc. Some arcs initialized with this procedure are visualized in Figure \ref{geometryinit}. Following initialization, a leaf region of interest (ROI) is defined based on the variability of leaf positions in the initialized arc (Figure \ref{leafvariability}). We measure the spread of positions with the standard deviation along the control-point axis and select the region specified by $\sigma_x > 2$ mm $\pm$ 2 leaves. Inverse planning with simulated annealing is then implemented in the ROI. At each iteration, 10\% of leaves in ROI are randomly selected and adjusted.

\section{Validation of Dose Calculations}

For machine simulation, we used the Python-based dose calculation software Conehead inspired by Cho \textit{et al.} and Yang \textit{et al.} \cite{peet2018conehead, cho2012practical, yang2002three}. Although this software fails to address some basic issues in dose calculation such as MLC leaf transmission, the beam model suffices to compare the VMAT optimization algorithms presented here. The quality of the dose calculations is not expected to favor one optimization scheme over the other.

\begin{figure}[!hbt]
\centering
\begin{subfigure}{0.32\textwidth}
  \centering
  \includegraphics[width=\linewidth]{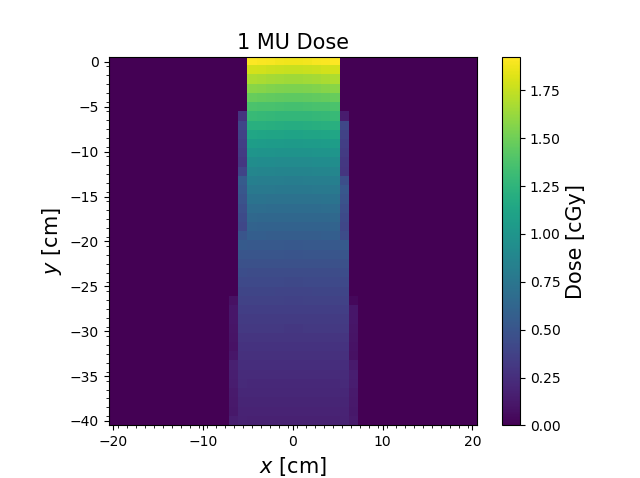}
\end{subfigure}
\begin{subfigure}{0.32\textwidth}
  \centering
  \includegraphics[width=\linewidth]{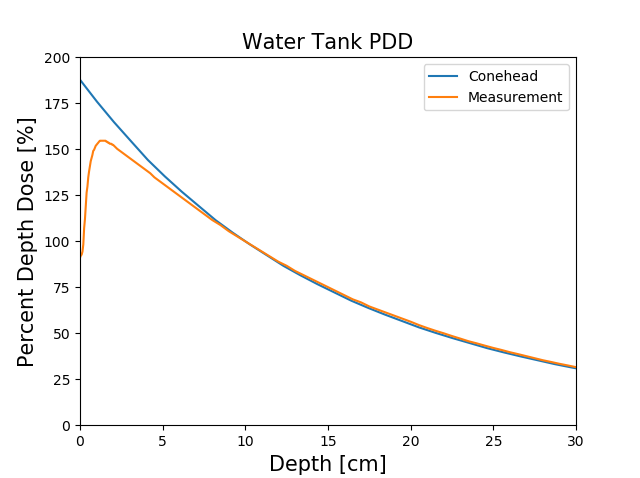}
\end{subfigure}
\begin{subfigure}{0.32\textwidth}
  \centering
  \includegraphics[width=\linewidth]{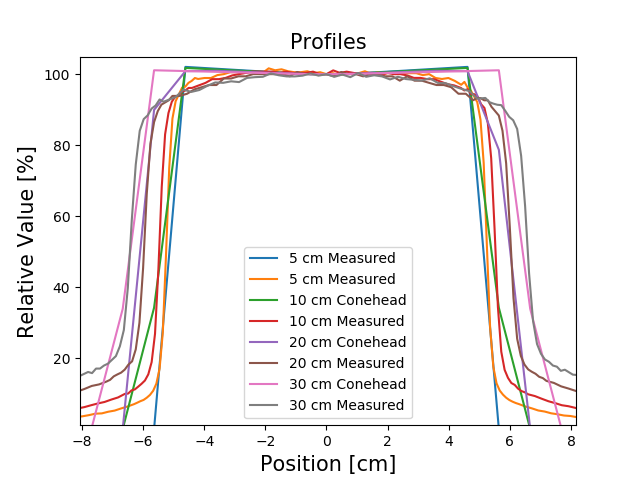}
\end{subfigure}
\caption[Water Tank Validation]{Dose deposition for an open 10 cm x 10 cm field in a water tank at 100 SSD. PDDs are normalized to dose at 10 cm depth, while profiles are normalized to central axis doses. While the PDD fails to model dose build up at shallower depths, Conehead is in high agreement with measured values beyond $\approx 5$ cm. While our dose calculation algorithm fails to capture beam softening at greater depths and large off axis distances, the overall shape of the beam profiles is fairly reproduced to within $\approx 10\%$.}
\label{watertank}
\end{figure}

In our experiments, we simulate a beam from a 6 MV Varian Clinac accelerator. The beam model was validated against clinically measured percent depth dose curves and beam profiles using a simulated water tank at 100 source to surface distance (SSD) (Figure \ref{watertank}).

While dose in the buildup region significantly disagrees with the measured PDD, doses at depths $> 5$ cm are in high agreement with the measured values. Similarly, we find that although the simulated beam exhibits some beam softening, off-axis doses at greater depths are typically overestimated, but are within 10\% at even 30 cm. These are not the highest possible accuracy dose calculations. They however suffice for the purpose of comparing VMAT optimizers.

\section{Gradient-based Latent Space Algorithms}\label{gradsection}

Gradient-based local DAO is a cornerstone for many clinical VMAT optimizers (SmartArc, Oncentra VMAT, RayArc, Monaco) \cite{unkelbach}. In local DAO with FMO-informed segment initialization, conventional convex IMRT optimization at individual gantry angles is used to generate high-quality arc therapy initializations. Gradients are then calculated and used to descend the optimizer state to a nearby local optimum (Figure \ref{traditionalalgs}). There may be some potential benefit to implementing gradient-based optimization for the latent space algorithms, either to quickly choose an initialization for a full gradient-based search or for complete optimization, depending on the quality of the result.

To do so, gradients of dosimetric objectives $f(d)$ with respect to the low-dimensional coordinates $z$ should be computable. Recall Equation \ref{dosesum}
\begin{equation*}
    d_k  = \sum_{i} D^\phi_{ik} x_{i}
\end{equation*}
which expresses the dose to voxel $k$ for beamlet $i$ at according to the dose influence matrix $D$ and apertures $x$. For the latent space algorithms, we have some generator $g$ which specifies a coordinate transformation $x = g(z)$, and we would like to compute
\begin{equation}
    \frac{\partial f}{\partial z_j} = \sum_{i, k} \frac{\partial f}{\partial d_k}\frac{\partial d_k}{\partial x_i}\frac{\partial x_i}{\partial z_j} \ .
\end{equation}
Here, the first term in the product will be easily computable for the user-specified objective function and second term is recognized as the dose influence matrix $\partial d_k /\partial x_i = D_{ik}$. What is left is the Jacobian $J_{ij} = \partial x_i / \partial z_j$ for the fully-parameterized and latent space coordinates, which will of course be specific to the learning algorithm,
\begin{equation}
    \frac{\partial f}{\partial z_j} = \sum_{i, k} \frac{\partial f}{\partial d_k} D_{ik} J_{ij} \ .
\end{equation}
In the case of direct aperture optimization, beamlets $x_i$ were specified according to Equation \ref{apdef}. The Jacobian for global DAO is easily computable as shown by Unkelbach \textit{et al.} in 2015 \cite{unkelbach}.  In principle, one can represent apertures however they want and perform gradient-based DAO so long as the Jacobian is computable.

\subsection{PCA Gradients}

In the case of PCA, the Jacobian is easily tractable. Beamlets $x_i$ can be represented as a linear combination of eigenarcs $\{v^{(1)}, v^{(2)}, ..., v^{(d)}\}$ with coefficients $z_j$,
\begin{equation}
    x_i = \sum_{j=1}^d z_j v_i^{(j)} \ .
\end{equation}
In this case, the Jacobian is simply the stacked eigenarcs
\begin{equation}
    J_{ij} = \frac{\partial x_i}{\partial z_j} = v_i^{(j)} \ .
\end{equation}

\subsection{AE Gradients}

The derivative of multi-layer neural network outputs with respect to their inputs can be computed according to the chain rule. Given hidden representations $h^{(1)}, h^{(2)}, ..., h^{(f)}$, the Jacobian can be written as a product of many partial derivatives
\begin{equation}\label{totalgradient}
    \frac{\partial x_i}{\partial z_j} = \sum_{k,l,m,n,...} \frac{\partial x_i}{\partial h^{(f)}_k} \frac{\partial h^{(f)}_k}{\partial h^{(f-1)}_l} \cdot ... \cdot \frac{\partial h^{(2)}_m}{\partial h^{(1)}_n} \frac{\partial h^{(1)}_n}{\partial z_j} \ .
\end{equation}

Our autoencoder generators are composed of five functions: matrix multiplication, transposed convolutions, ReLU activations, batch normalization, and dropout. Two of these functions, batch normalization and dropout, are turned off during model deployment. To compute the total gradient of the generator output with respect to its input, we must be able to calculate derivatives with respect to the inputs of linear, 2D convolution transpose, and ReLU layers. Derivatives with respect to the input of the remaining layers are easily computable as shown below.

\subsubsection{Linear Layers}

\noindent Linear layers modify the input with matrix multiplication,
\begin{equation}
    y_i = \sum_j M_{ij}x_j + B_i \ .
\end{equation}
The Jacobian for linear layers is simply the networks weights $M$,
\begin{equation}
    \frac{\partial y_i}{\partial x_j} = M_{ij} \ .
\end{equation}

\subsubsection{Transposed Convolution Layers}

The ordinary convolution operation used in convolutional neural networks can be expressed as matrix multiplication \cite{zhang2020dive}. For a $h_1$ x $w_1$ image $x$, we can perform 2D convolution with a $h_2$ x $w_2$ kernel $k$ by creating an intermediate matrix $K$ having shape $(h_2 w_2, h_1 w_1)$. We flatten $x$ and $k$ and organize $K$ such that every slice along the second axis corresponds to a mask of zeros except for the $h_2 w_2$ values specified by the kernel. In this setting, convolution is the matrix product
\begin{equation}\label{convseqn}
    y_i = \sum_{j=1}^{h_1 w_1} K_{ij} x_{j} \ .
\end{equation}
Striding and padding are added by simply adjusting the geometry of $K$.

Strided transposed convolutions are used in our networks to upsample image data. These are implemented similarly. To perform a transposed convolution from the output image space $Y$ to input image space $X$, we perform the same matrix arithmetic as shown in Equation \ref{convseqn}, except first taking the transpose of $K$,
\begin{equation}
    x_i = \sum_{j=1}^{h_2 w_2} K_{ji} y_{j} \ .
\end{equation}\label{convstransposeeqn}
In either case, convolution or transposed convolutions, the function can be represented as a matrix product having Jacobian
\begin{equation}
    \frac{\partial y_i}{\partial x_j} = K_{ij} \ .
\end{equation}

\subsubsection{ReLU Activations}

\noindent Rectified linear unit (ReLU) activations require that all outputs be non-negative $r(x_i) \geq 0$,
\begin{equation}
    r_i = \begin{cases}
        x_i & \text{if $x_i \geq 0$}\\
        0 & \text{otherwise}
    \end{cases} \ .
\end{equation}
The gradient in this case is trivial,
\begin{equation}
    \frac{\partial r_i}{\partial x_i} =
    \begin{cases}
        1 & \text{if $x_i \geq 0$}\\
        0 & \text{otherwise}
    \end{cases} \ .
\end{equation}

\subsection{Computing the Total Gradient}

With the gradients of the linear and transposed convolution layers expressed as matrices, we can begin to write the summation shown in Equation \ref{totalgradient} explicitly. Computing the Jacobian for an example block $h^{(i)} = B(h^{(i-1)}) = [$linear, ReLU, convolution, ReLU$]$ specified by matrices $M$ and $K$ will reduce to contracting those matrices,
\begin{gather}
    \sum_l M_{kl}K_{lm} = g_{km} \\
    \frac{\partial h^{(i)}_k}{\partial h^{(i-1)}_m} =
    \begin{cases}
        g_{km} & \text{if $ \geq 0$} \\
        0 & \text{otherwise}
    \end{cases} \ .
\end{gather}
The complete Jacobian for the whole network should be computable with a simple summation like this. 
\end{appendices}

\bibliographystyle{unsrt}  
\bibliography{references}

\end{document}